\documentclass[11pt,a4paper]{article}

\usepackage[T1]{fontenc}
\usepackage{amsmath,amssymb,amsthm}
\newcommand{\bigboxplus}{\mathop{\textstyle\sum\nolimits^{\boxplus}}}   %
\usepackage{booktabs,array}
\usepackage[margin=25mm]{geometry}
\usepackage[round]{natbib}
\usepackage{xcolor}
\usepackage[colorlinks=true,linkcolor=blue!60!black,citecolor=blue!60!black,urlcolor=blue!60!black]{hyperref}

\newcommand{\ThmName}{Theorem}          %
\newcommand{\PropName}{Proposition}     %
\newcommand{\LemName}{Lemma}            %
\newcommand{\DefName}{Definition}       %
\newcommand{\RemName}{Remark}           %
\newcommand{\ExaName}{Example}          %
\theoremstyle{plain}
\newtheorem{theorem}{\ThmName}[section]
\newtheorem{proposition}[theorem]{\PropName}
\newtheorem{lemma}[theorem]{\LemName}
\theoremstyle{definition}

\newtheorem{example}[theorem]{\ExaName}
\theoremstyle{remark}
\newtheorem{remark}[theorem]{\RemName}

\providecommand{\C}[1]{\mathcal{#1}}
\providecommand{\Set}[1]{\mathbb{#1}}
\providecommand{\mi}[1]{\mathit{#1}}

\newcommand{\HA}{\C{H}}

\newcommand{\ten}{\otimes}

\newcommand{\idm}{\mi{id}}
\newcommand{\cop}{\Delta}
\newcommand{\rcop}{\widetilde{\Delta}}
\newcommand{\cou}{\epsilon}              
\newcommand{\mult}{m}

\newcommand{\conv}{\star}

\newcommand{\SOz}{\C{SO}_0}
\newcommand{\Tset}{\mathfrak{T}_{\SOz}}
\newcommand{\Fset}{\mathfrak{F}_{\SOz}}

\newcommand{\Vsp}[1]{\C{V}(#1)}
\newcommand{\Accm}{\mi{Acc}}

\newcommand{\Mop}{\mathfrak{M}}          
\newcommand{\Bop}{\C{B}}                 

\newcommand{\RBalg}{\C{R}}               
\newcommand{\RBop}{R}                    

\newcommand{\Hsemi}{\HA^{\mi{semi}}}

\newcommand{\Bog}[1]{\widetilde{#1}}     

\newcommand{\tr}[1]{\langle #1\rangle}                      %
\newcommand{\HAb}{\overline{\HA}}                            %
\newcommand{\copb}{\overline{\cop}}                          %
\providecommand{\defequiv}{\mathrel{\stackrel{\mathit{def}}{\equiv}}}   %
\newcommand{\mcb}[1]{(MCB, #1)}                              %

\title{Does Merge Form a Hopf Algebra?\\ {\large Coproducts, Traces, and the Prospects of Algebraic Minimalism}}
\author{Daisuke Bekki\thanks{%
  This paper presents the author's analysis of Marcolli, Chomsky, and Berwick (2025), developed for, and revised after, the author's commentary at the workshop
  ``Computational Generative Linguistics: Overview and Prospect'' (Linguistic Society of Japan, 15 November 2026).
  A generative AI (Anthropic's Claude, via Claude Code) was used, under the author's direction, to draft the Japanese manuscript from the author's documents,
  to translate it into English, and to proofread it. The author read and checked every sentence, revised the claims, proofs, and citations, and verified all
  references against the original sources. The author is responsible for the entire content.}\\ Ochanomizu University}
\date{November 2026}   %

\begin{document}
\maketitle

\begin{abstract}

Marcolli, Chomsky, and Berwick (2025; henceforth MCB) formulate the Merge operation of Minimalism on a Hopf algebra of Connes--Kreimer type, and claim that language and physics obey the same algebraic structure.
This paper examines the premise of that claim: whether Merge on workspaces can be described as a Hopf algebra.
We first survey five coproducts for coassociativity and a connected grading: the Connes--Kreimer coproduct $\Delta_\rho$, MCB's two coproducts $\Delta_d$ and $\Delta_c$, the contraction coproduct of Marcolli and Skigin (2025), and the coproduct of Xiao (2026).
$\Delta_d$ is not coassociative. For $\Delta_c$, MCB gives no rule for the labels of trace leaves in trees that already contain trace leaves; under the literal extension it is not coassociative,
but once a rule and a grading are supplied it yields a graded connected Hopf algebra. The coproduct of Xiao is coassociative, and so is that of Marcolli and Skigin once its definition is extended to trees with trace leaves.
These properties do not tell the coassociative coproducts apart; what distinguishes them is the representation that movement leaves behind.
We then review how generative grammar has treated moved elements, and confirm that semantic interpretation requires two things of the representation that results from movement.
Retention requires that the structure the lower position had before movement be recoverable; Identifiability requires that, in interpreting the lower position, it be determinable which upper element, if any, corresponds to it.
None of the five coproducts meets both conditions: Retention is met only by $\Delta_c$, and there only in the form of a label, and Identifiability by none.
In light of the results on coassociativity, we also reassess the Birkhoff factorization that MCB employs at the syntax--semantics interface.
Finally, we indicate at least three directions for algebraic Minimalism.

\end{abstract}

\section{Introduction}
\label{sec:intro}

\citet{MCB2025} (henceforth MCB) is an attempt to re-implement the syntactic computation of Minimalism on top of a Hopf algebra.
It takes the Connes--Kreimer Hopf algebra \citep{CK1998} as its foundation:
syntactic objects are binary, nonplanar rooted trees with labeled leaves, and workspaces are forests of such trees.
Since these forests form a monoid, the vector space spanned by the set of workspaces can be constructed as a monoid algebra (so that the product is the juxtaposition of forests, $\sqcup$).
On top of this, a coproduct (the extraction of accessible terms) is introduced, and Merge is defined as an operator on this space.

The Connes--Kreimer Hopf algebra was originally developed as an algebra for quantum field theory.
The Birkhoff factorization that MCB employs at the syntax--semantics interface is likewise a mechanism used in the renormalization of quantum field theory.
The intention behind MCB's program is stated plainly in the introduction.

\begin{quote}
\small
``It is in our view a striking result of significant importance to explicitly discover that, within the setting of generative linguistics, one can encounter the same level of highly structured
coherence governed by precise algebraic conditions, as in modern physics. Not only that, but in fact
the algebraic structures involved are exactly those that govern similar phenomena within the context of
theoretical physics.'' \mcb{p.~4}
\end{quote}

\noindent In the surrounding passages Chomsky's ``Miracle Creed,'' Galileo, and Feynman are invoked, and on p.~94 the point is pressed further:
\begin{quote}
``the first observation identifies the generative process of syntactic objects through Merge with the basic case of
the structure of generative processes of fundamental physics. The second observation suggests that the optimality that the core
computational structure of Merge ought to satisfy has the same conceptual nature as the least action principle of physics,
when the latter manifests itself in combinatorial form'' \mcb{p.~94}
\end{quote}

\noindent The section titles, too, include \S1.17 ``Merge and Fundamental Combinatorial Recursions in Physics'' and
\S2.5.3 ``Physics Methods in Minimalism.''
MCB's program, in other words, is not merely to rewrite Minimalism in algebraic terms,
but to claim that language and physics share the same algebraic structure.

This paper examines the program by separating it into two parts.
\begin{description}
\item[Working hypothesis H] Minimalism (Merge acting on workspaces) can be described as a Hopf algebra of Connes--Kreimer type.
\item[Claim P] If H holds, then natural language and quantum field theory obey the same principle.\footnotemark
\end{description}%
\footnotetext{%
To assert P, it does not suffice to show that Minimalism and quantum field theory share the same algebraic structure.
Hopf algebras on trees are not a device peculiar to renormalization: they appear in many fields, including Butcher's composition group of numerical integration methods \citep{Butcher1972},
the Hopf algebra of planar binary trees \citep{LR1998}, semiring parsing \citep{Goodman1999}, and operads.
What it would take for sharing a structure to count as ``obeying the same principle'' is not addressed here and is left for another occasion. This paper confines itself to the examination of H.
}%
H is a mathematically verifiable proposition, whereas P is an interpretation placed on top of it.
This paper mainly examines H and argues the following.
\begin{enumerate}
\item MCB has two coproducts, and they can be regarded as two different ``translations'' of movement (Internal Merge) in Minimalism.
      The translation that cancels the lower copy is $\cop_d$, the one that leaves what MCB calls a trace%
\footnote{%
Two things are called traces in this paper. One is the trace of generative grammar up to \textit{Barriers} \citep{Chomsky1986}, and of the generative grammar presupposed by \citet{HK1998}: an indexed element that remains in the position from which movement started.
The other is the trace of algebraic Minimalism: the label $\tr{T_v}$ that MCB assigns to a leaf of the contraction quotient $T/_c T_v$ \mcb{Def.~1.2.4, p.~30}. In what follows, the bare word ``trace'' refers to the former; a leaf carrying MCB's label is called a \emph{trace leaf}, and $\tr{T_v}$ its label.
MCB writes this label as $T_v$ struck through; in this paper we write $\tr{T_v}$ instead.}
is $\cop_c$, and MCB uses the former on the syntactic side and the latter on the semantic side
      (\S\ref{sec:two-coproducts}).
\item We examine coassociativity and connected gradings for five coproducts: the Connes--Kreimer coproduct $\cop_\rho$, MCB's two coproducts $\cop_d$ and $\cop_c$, the contraction coproduct $\cop_c^{h}$ of \citet{MarcolliSkigin2025}, and the coproduct of \citet{Xiao2026}.
      $\cop_d$ is not coassociative, and hence not even a coalgebra.
      As for $\cop_c$, MCB does not specify the label of a trace leaf when a tree that already contains trace leaves is contracted; under the literal extension it is not coassociative,
      but we show that, once a rule is supplied under which nested contractions leave the labels unchanged and the grading is chosen anew, it becomes a graded connected Hopf algebra.
      The coproduct of Xiao has already been shown to be coassociative \citep[Theorem~2.1]{Xiao2026}; that of Marcolli and Skigin is coassociative once its definition is extended to trees with trace leaves (\S\ref{sec:not-hopf}).
\item As regards the specification of Internal Merge, generative grammar has moved from ``movement that leaves an indexed trace'' to the conception on which a copy is left, Merge applies, and the copy is deleted at Externalization.
      The semantic interpretation required by standard formal semantics, however, demands two things of the representation that results from movement: \emph{Retention}, that the structure the lower position had before movement be recoverable,
      and \emph{Identifiability}, that, in interpreting the lower position, it be determinable which upper element, if any, corresponds to it.
      None of the five coproducts meets both conditions: Retention is met only by $\cop_c$, and there only in the form of a label, and Identifiability by none (\S\ref{sec:binding}).
\item In the setting of MCB's own text, the Birkhoff factorization is not derived as a theorem about Hopf algebras; it is given only by a direct construction.
      Under $\cop_c$ with the rule for the labels of trace leaves and the grading introduced in \S\ref{sec:not-hopf}, and under the coproduct of \citet{Xiao2026}, the premises of the theorem are satisfied (\S\ref{sec:birkhoff}).
\item Coassociativity can thus be repaired, whereas a problem remains with the representation that a coproduct leaves as the result of movement.
      We indicate at least three directions in which algebraic Minimalism may proceed from here: not to require Retention and Identifiability, to design a coproduct that satisfies them, and not to require a Hopf algebra (\S\ref{sec:direction}).
\end{enumerate}

We assume that the reader is familiar with the basics of Minimalism (Merge, workspaces, the copy theory of movement) and with the definition of a Hopf algebra
(bialgebra, coassociativity, graded connectedness, antipode; see, e.g., \citealp{Sweedler1969}).
Quotations from MCB are located by section number and printed page.

\section{MCB's Merge}
\label{sec:merge}

We begin by reviewing MCB's definition of Merge, together with what each of its stages does linguistically.

\paragraph{The space of workspaces.}
A syntactic object is a binary, nonplanar rooted tree whose leaves are labeled by elements of the set $\SOz$ of lexical items (and syntactic features) \mcb{Def.~1.1.1, Remark 1.1.2, pp.~22--23}.%
\footnote{
MCB also introduces, by means of the FormSet operation, structures whose root has more than two daughters \mcb{\S1.16, pp.~141--147},
but sets them aside in the rest of the book, returning to them only in \S3.8.2 \mcb{p.~147}. We follow MCB in this and consider binary trees only.
}
We write $\Tset$ for the set of syntactic objects.
A workspace is a nonempty finite multiset of syntactic objects, that is, a forest \mcb{Def.~1.2.1, p.~27}.
The set of all forests together with the empty forest $1$ is a commutative monoid with the juxtaposition of forests $\sqcup$ as product and $1$ as unit \mcb{p.~29}; it is nothing other than the free commutative monoid generated by $\Tset$ (we write $\Fset$ for it).
The space of workspaces is the $\Set{Q}$-linear space with basis $\Fset$,
\[
\HA \defequiv \Vsp{\Fset}
\]
equipped with the product and the unit of the monoid, that is, the $\Set{Q}$-monoid algebra $\Set{Q}[\Fset]$ of $\Fset$ \mcb{p.~29}.

\paragraph{Accessible terms.}
For a syntactic object $T$, let $V_{\mi{int}}(T)$ be the set of non-root vertices of $T$ (leaves included),%
\footnote{MCB calls the elements of $V_{\mi{int}}(T)$ the ``internal (non-root) vertices,'' but
(1.2.3), $\mi{Acc}'(T) = \{T\} \cup \mi{Acc}(T)$, and (1.2.5), $\#V(F) = b_0(F) + \#\mi{Acc}(F)$, show that
all vertices other than the root, leaves included, are meant. Since leaves count as accessible terms, a single lexical item can also be an argument of Merge.}
and for $v \in V_{\mi{int}}(T)$ write $T_v$ for the subtree consisting of $v$ and its descendants.
The set of accessible terms of $T$ is $\Accm(T) = \{\,T_v \mid v \in V_{\mi{int}}(T)\,\}$ \mcb{Def.~1.2.2, p.~28}.
In Minimalist terms, this is the set of constituents that can serve as arguments of Merge, that is, the set of syntactic objects that are proper parts of $T$.
We adopt the convention that trace leaves (\S\ref{sec:two-coproducts}) are not counted among the accessible terms; MCB's Def.~1.2.2 is not explicit on this point.
The convention matters: without it, the terms in which a trace leaf is itself extracted have no counterpart in the proof of coassociativity, so that $\cop_c$ would fail to be coassociative under every rule for the labels of trace leaves considered in \S\ref{sec:delta-c}, even the label-forgetting one;
Propositions~\ref{prop:resolved} and \ref{prop:grading} also rely on it.

\paragraph{Coproduct.}
MCB's coproduct is the linear map that enumerates every way of extracting a collection of accessible terms from a workspace.
For a syntactic object $T$ and a collection $F_v = T_{v_1} \sqcup \cdots \sqcup T_{v_n}$ of pairwise disjoint accessible terms,
one defines ``the remainder after extracting $F_v$,'' $T/_\omega F_v$ (there are three ways of doing so, $\omega \in \{c,\rho,d\}$; see \S\ref{sec:two-coproducts}), and sets
\begin{equation}
\label{eq:coproduct}
\cop_\omega(T) \defequiv T \ten 1 + 1 \ten T + \sum_{F_v} F_v \ten T/_\omega F_v
\end{equation}
\mcb{Def.~1.2.8, (1.2.8), p.~33}.
The left factor is ``what is extracted,'' and the right factor is ``the workspace after the extraction.''%
\footnote{The term $T \ten 1$ extracts the whole tree $T$, regarded as an accessible term of the workspace
\mcb{(1.2.9), p.~33}; by the multiplicative extension, the coproduct of a forest also contains terms that extract whole components (trees).
For instance, for $F = \mi{john} \sqcup \mi{loves} \sqcup \mi{mary}$
the expansion of $\cop_\omega(F)$ contains $(\mi{loves} \sqcup \mi{mary}) \ten \mi{john}$, and External Merge uses this term.
The accessible terms $\Accm(T)$ of a single tree do not include the root ($T$ itself), but in the coproduct of a workspace it is $\Accm'(T) = \{T\} \cup \Accm(T)$ that is operative.}

MCB extends this coproduct to forests by
\[
\cop_\omega(F \sqcup F') = \cop_\omega(F)\,\cop_\omega(F')
\]
\mcb{p.~33}; that this definition is well defined and compatible with the product $\sqcup$ can be seen from the definition of forests, as follows.
$\Fset$ is the free commutative monoid generated by $\Tset$, and $(\HA \ten \HA, \cdot, 1_{\HA \ten \HA})$ is a commutative monoid (since $\sqcup$ is commutative).
Hence, by the universal property of free commutative monoids, $\cop_\omega$ as defined on $\Tset$ determines a unique monoid homomorphism from $\Fset$.
Further, by the universal property of monoid algebras, this monoid homomorphism extends uniquely to an algebra morphism from the algebra $\Set{Q}[\Fset]$ to the algebra $\HA\ten\HA$.
The counit $\cou : \HA \to \Set{Q}$ sends the empty forest $1$ to $1$ and every nonempty forest to $0$.

\paragraph{The action of Merge.}
For each pair $S, S'$ of syntactic objects, MCB defines an operator $\Mop_{S,S'} : \HA \to \HA$ on workspaces by
\begin{equation}
\label{eq:merge}
\Mop_{S,S'}(F) \defequiv \sqcup \circ (\Bop \ten \idm) \circ \delta_{S,S'} \circ \cop_d(F)
\end{equation}
\mcb{Def.~1.3.4, (1.3.7), pp.~42--43}. Here $\delta_{S,S'}$ is the projection that keeps, among the terms of $\cop_d(F)$,
only those whose left factor coincides with $S \sqcup S'$ \mcb{Def.~1.3.1, p.~41}; it acts only on the left channel of the coproduct \mcb{p.~44}.
$\Bop$ is the grafting operator, which creates a new root above the two roots of the forest $S \sqcup S'$ and returns the tree $\{S,S'\}$
\mcb{Def.~1.3.2, p.~41}. This is precisely Merge as so-called ``set formation,'' $\Mop(S,S') = \{S,S'\}$.
The final $\sqcup$ juxtaposes the resulting tree with the right factor (the remaining workspace).
Written in terms of the types of the maps, this is the following chain.
\begin{equation}
\label{eq:merge-chain}
\HA \xrightarrow{\ \cop_d\ } \HA \ten \HA
\xrightarrow{\ \delta_{S,S'}\ } \HA \ten \HA
\xrightarrow{\ \Bop \ten \idm\ } \HA \ten \HA
\xrightarrow{\ \sqcup\ } \HA
\end{equation}

Reading each stage linguistically yields Table~\ref{tab:merge-stages}.
\begin{table}[t]
\centering
\small
\begin{tabular}{@{}p{0.17\textwidth}p{0.29\textwidth}p{0.48\textwidth}@{}}
\toprule
Stage & Mathematics & What it does linguistically \\
\midrule
$\cop_d$ (coproduct) & $F \mapsto \sum_{F_v} F_v \ten F/_d F_v$ &
  Enumerates all combinations of accessible terms that can be extracted from the workspace. The right factor is the workspace remaining after their extraction, with the lower copies canceled \\
$\delta_{S,S'}$ (projection) & Keeps only the terms whose left factor is $S \sqcup S'$ &
  Selects the two arguments of Merge. The choice of $S,S'$ is supplied from outside this formalization \\
$\Bop$ (grafting) & $S \sqcup S' \mapsto \{S,S'\}$ &
  Merge itself. Always creates a new root. \\
$\sqcup$ (product) & $\{S,S'\} \sqcup (\text{remainder})$ &
  Returns the newly formed syntactic object to the remaining workspace \\
\bottomrule
\end{tabular}
\caption{The four stages of MCB's Merge \eqref{eq:merge} and their linguistic reading.}
\label{tab:merge-stages}
\end{table}

In MCB, the differences among External Merge, Internal Merge, and Sideward Merge reduce to differences in the terms picked up by $\delta_{S,S'}$.
If $S$ and $S'$ are distinct components of the workspace, the result is External Merge.
Internal Merge is realized as the two-step composition $\Mop_{T/\beta,\beta} \circ \Mop_{\beta,1}$:
first $\Mop_{\beta,1}$ extracts an accessible term $\beta$ of $T$ to form the two components $\beta \sqcup T/\beta$, and then $\Mop_{T/\beta,\beta}$ performs External Merge on them
\mcb{Prop.~1.4.2, p.~50}.
MCB shows as a lemma that \eqref{eq:merge} agrees with the action of Merge as described by Chomsky and his colleagues (as long as the corresponding term exists)
\mcb{Lemma 1.3.6, p.~44}.

In this definition, the only place where the Hopf algebra is involved is $\cop_d$. The maps $\delta$, $\Bop$, and $\sqcup$ are operations on forests, extended linearly as they stand;
what carries algebraic content is the writing of ``the enumeration of Merge candidates'' as a coproduct.
We turn to this coproduct in the next section.

\section{MCB's Two Coproducts}
\label{sec:two-coproducts}

The ``remainder'' $T/F_v$ in the coproduct \eqref{eq:coproduct} can be defined in three ways. MCB's own account is as follows \mcb{pp.~30--33}.

\begin{description}
\item[Contraction quotient $T/_c F_v$] Each $T_{v_i}$ is contracted to its root $v_i$. The vertex $v_i$ becomes a leaf carrying the label $\tr{T_{v_i}}$.
  MCB calls this label the trace of $T_{v_i}$ \mcb{Def.~1.2.4, p.~30}; we call the leaf a trace leaf.
\item[Admissible-cut quotient $T/_\rho F_v$] The edge above each $T_{v_i}$ is cut (an admissible cut). The upper vertex of each cut edge remains as a nonbranching vertex with a single child,
  so the result is no longer a binary tree
  \mcb{Def.~1.2.6, p.~31; Fig.~1.2, p.~32}.
\item[Deletion quotient $T/_d F_v$] Each $T_{v_i}$ is removed from $T$, and one takes the unique maximal binary tree obtained from what remains by contracting edges
  \mcb{Def.~1.2.5, p.~31}.
\end{description}

\begin{example}
\label{exa:three-quotients}
Extracting $\beta$ from $T = \{\alpha, \{\beta,\gamma\}\}$ gives
$T/_c\beta = \{\alpha,\{\tr{\beta},\gamma\}\}$;
$T/_\rho\beta$ is the tree in which the root of $\{\beta,\gamma\}$ remains as a nonbranching vertex with the single child $\gamma$ (we write it $\{\alpha,(\gamma)\}$);
and $T/_d\beta = \{\alpha,\gamma\}$.
\end{example}

Of the three, $\cop_\rho$, defined by the admissible-cut quotient, is exactly the coproduct of the Connes--Kreimer Hopf algebra of rooted trees \mcb{Remark 1.2.9, p.~34}.
It is coassociative and makes $V(\tilde{\mathfrak F}_{\SOz})$ a graded connected Hopf algebra, graded by the number of vertices as in Connes--Kreimer \mcb{Lemma 1.2.11, pp.~38--39}.
As a representation of movement in Minimalism, however, $\cop_\rho$ is mismatched in two respects.
First, the tree that remains after movement is no longer binary;
second, the structure of the moved constituent cannot be recovered from the unlabeled vertex that it leaves behind.

As for the latter, in Chomsky's copy theory, Internal Merge leaves the original syntactic object unchanged (the No Tampering Condition), so that the moved element appears in two positions \citep[ch.~3]{Chomsky1995}.
The lower copy reaches the conceptual-intentional interface with its full structure intact, and that it goes unpronounced is determined at Externalization \citep{ChomskyEtAl2023}.
Neither traces nor indices are used.\footnote{The word itself survives: \citet[p.~30]{Chomsky2024GE} writes,
``\textit{The familiar notions trace and PRO remain, but with a principled formulation that follows from SMT. They are the two possible cases of Copy}.''
There, ``trace'' is a name for one kind of copy, not a separate indexed element.}
Chomsky and his colleagues state that the restriction to simplest Merge entails an Inclusiveness Condition that
``\textit{precludes the introduction of extraneous objects---for instance, traces and the bar-levels of X-bar Theory and other labels, but not copies}''
\citep[p.~237]{ChomskyGallegoOtt2019}, and go so far as to call the treatment in terms of traces and coindexing one ``\textit{in radical violation of IC}'' \citep[p.~246, n.~22]{ChomskyGallegoOtt2019}.
Reconstruction is taken to be a consequence of the lower copy being used for interpretation as it stands:
``\textit{All copies are available at SEM, rendering reconstruction operations of earlier theories obsolete}'' \citep[p.~236]{ChomskyGallegoOtt2019}.

MCB presents its formulation as replicating this copy theory (``\textit{required to replicate the copy theory of movement}'' \mcb{p.~10}), but the two are not the same.
In MCB, the extracted $T_v$ is placed in the left channel of the coproduct, and simultaneously the lower copy is canceled in the right channel:
``\textit{we never actually see any copy},'' and the resulting workspace is one ``\textit{containing possible repetitions but no copies}'' \mcb{\S1.8, p.~83}.
The lower copy survives only as the label of a leaf in $T/_c$ \mcb{p.~33}, and nothing of it survives in $T/_d$.
As \citet{Collins2026} points out, this violates the No Tampering Condition, and since only a vertex without structure remains in the lower position,
reconstruction for binding and scope cannot be accounted for in the way the copy theory does.
On the phonetic side the effects of the two coincide (the lower copy is not pronounced), but on the semantic side they do not.

$\cop_c$ is not a coproduct that MCB defines on its own; MCB presents it as the restriction to binary trees of the coproduct of the Connes--Kreimer Hopf algebra of Feynman graphs
(as \S\ref{sec:delta-c} shows, whether the version that retains the labels of trace leaves inherits coassociativity from this restriction depends on the rule for labels on trees that already contain trace leaves).
It also has what MCB claims corresponds to traces: the contracted vertex keeps the label of the subtree (the lower copy of the copy theory).
In this respect $\cop_c$ partly contains a solution to the second mismatch mentioned above: the leaf that $\cop_c$ leaves behind retains only the label of the moved element. Which of the elements above is the one that moved from that position, however, cannot be identified from this label (\S\ref{sec:binding}).

$\cop_d$ is a coproduct of MCB's own: it deletes the copy at the launching site without leaving anything behind, and leaves a binary tree. In this respect it resolves the first mismatch, but the second remains.

These two coproducts are, so to speak, MCB's two different translations of movement in Minimalism.
MCB chooses to assign these two coproducts to the syntactic side and the semantic side \mcb{p.~33; Def.~1.3.4, p.~42}:
\begin{itemize}
\item Syntactic side: the definition of Merge \eqref{eq:merge} uses $\cop_d$ \mcb{Def.~1.3.4, p.~42}.
  Externalization, too, uses the deletion quotient $T/_d$.%
  \footnote{
  ``\textit{the two different forms $T/_cT_v$ and $T/_dT_v$ of quotient account for the fact that the deeper copy of the accessible term $T_v$ is interpreted at the syntax--semantics interface but not expressed in Externalization}'' \mcb{\S3.1.2.1, p.~206}.
  }%
\item Semantic side: at the syntax--semantics interface it uses $\cop_c$, which leaves trace leaves \mcb{p.~206}.%
  \footnote{
  ``\textit{when it comes to interfacing syntax with semantics, it is generally better to retain the root vertex $v$ of $T_v$
  in the quotient $T/_cT_v$ with the $T_v$ label}''
  \mcb{\S3.1.2.1, p.~206}.
  }%
\end{itemize}

$\cop_c$ and $\cop_d$ live on the same space $\HA$, and are related, via $\cop_\rho$, by projections on the right channel \mcb{(1.3.10), p.~44}:
\begin{equation}
\label{eq:projection}
\cop_d = (\idm \ten \Pi_{d,\rho}) \circ \cop_\rho = (\idm \ten \Pi_{d,c}) \circ \cop_c
\end{equation}
Since $\delta_{S,S'}$ acts only on the left channel,
the syntactic object newly created by Merge is the same whichever coproduct is used; what differs is only the shape of the remaining workspace.

Now, MCB holds that the cancellation of copies is algebraically motivated.
That is, cancellation is
``\textit{necessary to have a `good,' well-behaved, coassociative coproduct $\Delta_c$},''
and coassociativity fails unless the copy of $T_v$ remaining in the right factor is quotiented out \mcb{\S1.8, p.~81}.
This argument, however, establishes only a necessary condition: a coproduct $T \mapsto \sum_v T_v \ten T$ that leaves $T$ intact in the right factor fails to be coassociative for every form of the quotient.
It does not show that coassociativity holds for $\cop_d$,
and the coproduct it names as coassociative is $\cop_c$.
Indeed, whereas Def.~1.3.4 states explicitly that $\Delta = \Delta_d$, the opening of \S1.8, which explains the same action, describes cancellation in terms of $T/_c$
and treats $T/_d$ as the case of Externalization \mcb{p.~81}.
The coassociativity of $\cop_c$ is thus used to motivate the deletion performed by $\cop_d$, and the argument is confused at this point.
In fact, coassociativity fails for $\cop_d$, and $(\HA,\sqcup,\cop_d)$, which MCB uses to implement Merge, is not a Hopf algebra; it is not even a coalgebra.
We discuss this in the next section.

Whether Merge can be written with the Connes--Kreimer apparatus is the touchstone of the working hypothesis H.
The assembling half of the action of Merge, the grafting $\Bop$, is realized within the Connes--Kreimer algebra (it corresponds to $B_+$ and satisfies the Hochschild 1-cocycle condition \mcb{\S1.17, p.~149}).
What is at issue is the extracting half, namely whether the specification of Merge, including what movement leaves in the lower position, can be expressed by a coassociative coproduct; the next two sections examine this.

\section{Five Coproducts and Coassociativity}
\label{sec:not-hopf}

The following two properties are required for an algebra equipped with a coproduct to be a graded connected Hopf algebra. In this section we show that MCB's two coproducts do not, as they stand, satisfy them.
\begin{itemize}
\item the coassociativity of the coproduct: $(\cop \ten \idm)\circ\cop = (\idm \ten \cop)\circ\cop$ (that is, ``two successive extractions commute'').  
Without coassociativity, the convolution product $f \conv g = \mult \circ (f \ten g) \circ \cop$ is not associative in general,
characters do not form a group, and neither the antipode (the convolution inverse of $\idm$) nor the theory of Birkhoff factorization is available.

\item the connectedness of a graded bialgebra: that is, the degree-$0$ part consists of the scalars only.  
This is presupposed by the existence of the antipode (Takeuchi's theorem \citep{Takeuchi1971}) and by the Birkhoff factorization theorem.
\end{itemize}

In connection with these problems, further coproducts have been put forward, besides MCB's two coproducts $\cop_d$ and $\cop_c$, by \citet{MarcolliSkigin2025} and by \citet{Xiao2026}.
We review these four, together with the original Connes--Kreimer coproduct $\cop_\rho$, with respect to the two properties above.
The results are summarized in Table~\ref{tab:four-coproducts}.

\subsection{$\cop_\rho$: Admissible Cuts}
\label{sec:delta-rho}
$\cop_\rho$ is exactly the coproduct of the Connes--Kreimer Hopf algebra of rooted trees \mcb{Remark 1.2.9, p.~34}. It is coassociative, and with the grading by the number of vertices it yields a connected Hopf algebra \mcb{Lemma 1.2.11, pp.~38--39}.
The trees in the right channel, however, do not stay within binary trees (the upper vertex of each cut edge remains as a vertex with a single child). MCB uses this coproduct as an intermediate form and does not use it in the definition of Merge.

\subsection{$\cop_d$: Deletion}
\label{sec:delta-d}

As MCB itself acknowledges, the coproduct $\cop_d$ defined by the deletion quotient is not coassociative.%
\footnote{%
MCB states that $\cop_d$ satisfies only ``\textit{a weaker relation where the lists of terms for the coassociativity relation match, but only up to
certain multiplicities in some of the terms},''
and defers the correction of multiplicities to a separate paper (MCB's reference [146], Marcolli--Walton, ``Generalized Quasi-Hopf Algebras and Merge'')
\mcb{Remark 1.2.9, p.~34; Lemma 1.2.12, p.~39}.
That paper has not appeared as of this writing.
We note that the failure of coassociativity of $\cop_d$ itself is also stated explicitly by \citet[\S1.4.2]{Xiao2026}:
``\textit{The deletion coproduct is not coassociative: although $(\Delta^d \otimes 1)\circ\Delta^d$ and $(1 \otimes \Delta^d)\circ\Delta^d$ enumerate the same terms, they may appear with different multiplicity.}''
Proposition~\ref{prop:delta-d} is therefore not a new result; it makes the minimal counterexample explicit.%
}
The minimal counterexample is as follows.

\begin{proposition}
\label{prop:delta-d}
Let $T = \{\alpha,\{\beta,\gamma\}\}$. The coefficients of $\beta \ten \gamma \ten \alpha$ in $(\cop_d \ten \idm)\circ\cop_d(T)$ and in $(\idm \ten \cop_d)\circ\cop_d(T)$
are $2$ and $1$, respectively. Hence $(\HA, \cop_d)$ is not a coalgebra.
\end{proposition}

\begin{proof}
On the left-hand side, $\beta \ten \gamma \ten \alpha$ arises from those terms $X \ten \alpha$ of $\cop_d(T)$ for which $\cop_d(X)$ contains $\beta \ten \gamma$.
There are two such $X$, namely $X = \{\beta,\gamma\}$ and $X = \beta \sqcup \gamma$, and both have the deletion quotient $T/_d X = \alpha$
(whether $\{\beta,\gamma\}$ is deleted, or $\beta$ and $\gamma$ are deleted separately, the maximal binary tree that remains is $\alpha$).
$\cop_d(\{\beta,\gamma\})$ contains $\beta \ten \gamma$ once, and so does
$\cop_d(\beta \sqcup \gamma) = \cop_d(\beta)\,\cop_d(\gamma) = (\beta \ten 1 + 1 \ten \beta)(\gamma \ten 1 + 1 \ten \gamma)$.
In total, $2$.
On the right-hand side, the same term arises from the terms $\beta \ten Y$ of $\cop_d(T)$ for which $\cop_d(Y)$ contains $\gamma \ten \alpha$,
and there is only one such $Y$, namely $Y = T/_d\beta = \{\alpha,\gamma\}$. Since $\cop_d(\{\alpha,\gamma\})$ contains $\gamma \ten \alpha$ once, the coefficient is $1$.
\end{proof}

The cause is that the distinct subforests $\{\beta,\gamma\}$ and $\beta \sqcup \gamma$, that is, distinct derivational paths, collapse onto the same remainder $\alpha$ and lose their distinction.

MCB announces a separate paper that handles this discrepancy by correcting the multiplicities. One remark is in order on the quasi-Hopf algebras named in its title.
A quasi-Hopf algebra \citep{Drinfeld1990} is a structure in which coassociativity is relaxed only up to conjugation by an invertible element $\Phi \in \HA^{\ten 3}$:
\[
(\idm \ten \cop) \circ \cop(x) = \Phi \,\bigl((\cop \ten \idm) \circ \cop(x)\bigr)\, \Phi^{-1}.
\]
The slack is conjugation and nothing more. But the algebra $\HA$ of workspaces is commutative under $\sqcup$, so $\HA^{\ten 3}$ is commutative as well, and conjugation is the identity map.

\begin{proposition}
\label{prop:quasi-hopf}
A Drinfeld quasi-bialgebra whose underlying algebra is commutative is strictly coassociative.
Consequently, since $\cop_d$ is not coassociative (Proposition~\ref{prop:delta-d}), $(\HA, \sqcup, \cop_d)$ is not a quasi-bialgebra in Drinfeld's sense for any choice of $\Phi$,
and hence cannot be a quasi-Hopf algebra either.
\end{proposition}

The proof is given in Appendix~\ref{sec:proofs-quasi-hopf} (it takes two lines).
This is presumably why the title of MCB's separate paper reads ``\textit{Generalized} Quasi-Hopf Algebras and Merge.''

\subsection{Marcolli and Skigin's Coproduct: Head Labels}
\label{sec:ms}
\citet{MarcolliSkigin2025}, a paper following MCB, refines the definition of the labels of the trace leaves of the contraction coproduct $\cop_c$, presenting this as a clarification of MCB's account (\S10.1, pp.~47--48).
In MCB's definition \mcb{Def.~1.2.4, p.~30}, the label records the contracted subtree $T_v$ itself, but the set of labels then becomes countably infinite.
The paper holds that the label need not retain the whole structure of $T_v$:
``\textit{After all, the extracted term $T_v$ is still present, on the other side of the coproduct, so that information is not lost}''
(p.~47).
Accordingly, assuming that each tree $T$ is equipped with a head function $h_T$, it defines the label of the trace leaf arising from the contraction of $T_v$
to be the lexical item of the head leaf $h_T(v)$ of $T_v$, marked with a cancellation symbol (Definitions 10.1--10.3).
The set of labels is a copy of $\SOz$, and is finite when $\SOz$ is. In what follows we write $\cop_c^{h}$ for the contraction coproduct with this rule.

The paper states that coassociativity holds under this rule (Lemma 10.4).
This can also be checked by the condition of Proposition~\ref{lem:label-invariance} in \S\ref{sec:delta-c}: whatever has already been extracted from inside a subtree $X$, the lexical item of the head of $X$ does not change.
The range over which the label is defined, however, raises the same problem as for MCB's $\cop_c$, and the statement holds only under the reading supplied in \S\ref{sec:delta-c},
which includes the convention that trace leaves are not counted among the accessible terms.
As for the grading, under the same reading the argument of Proposition~\ref{prop:grading} does not depend on the labels of trace leaves, and so applies to $\cop_c^{h}$ as it stands.

\subsection{Xiao's Coproduct: Restricting the Cuts}
\label{sec:xiao}
The problem of the coassociativity of $\cop_d$ itself has been solved by \citet{Xiao2026} in a different way, namely by restricting the cuts that enter the sum.
What makes the coefficients disagree in the counterexample of Proposition~\ref{prop:delta-d} is the subforest $\beta \sqcup \gamma$, which extracts both children $\beta$, $\gamma$ of one and the same vertex.
\citet{Xiao2026} defines a coproduct whose sum ranges only over those Connes--Kreimer admissible cuts that do not cut both children of the same vertex (binary-admissible cuts).
The right channel is, as with $\cop_d$, the remainder reduced to a binary tree. This coproduct is shown to be coassociative (Theorem 2.1 there), and its terms correspond one-to-one to the subsets of the leaves (Proposition 2.1 there).
The space is the monoid algebra of the forests of nonplanar binary trees, and the coproduct is extended from trees to forests multiplicatively, so its compatibility with the product $\sqcup$ holds by construction. It is connected under the grading by the number of leaves, and is therefore a Hopf algebra.
Here the problem of the coassociativity of $\cop_d$ is resolved, by restricting the cuts that enter the sum, while binary trees and nonplanarity are retained.

\subsection{$\cop_c$: Contraction and the Labels of Trace Leaves}
\label{sec:delta-c}

As we saw in \S\ref{sec:two-coproducts}, MCB takes $\cop_c$ to be the restriction of the Connes--Kreimer coproduct, and hence coassociative. %
MCB's text, however, does not state under what assumptions this coassociativity holds.
MCB's definition \mcb{Def.~1.2.4, p.~30} specifies the label $\tr{T_v}$ of a trace leaf only for a syntactic object $T \in \mathfrak{T}_{\SOz}$ that contains no trace leaves and an accessible term $T_v$ of it.
The right-hand side of the coassociativity law $(\cop_c \ten \idm)\circ\cop_c = (\idm \ten \cop_c)\circ\cop_c$ applies $\cop_c$ once more to the tree $T/_cF$, which already contains trace leaves,
so the label must also be specified for the contraction of a subtree that contains trace leaves.
MCB gives no such rule, and as the ground for coassociativity merely points out that $\cop_c$ is the restriction of the Connes--Kreimer Hopf algebra of Feynman graphs \mcb{Remark 1.2.9, p.~34}.
\citet[\S1.4.1]{Xiao2026} likewise states the same ground in a single sentence.
On the Connes--Kreimer side, however, a contracted vertex does not carry the content of the subgraph as a label, so what this ground directly supports is only the label-forgetting version.

The same applies to the coproduct $\cop_c^{h}$ of \citet{MarcolliSkigin2025} (\S\ref{sec:ms}).
That paper replaces the label by the lexical item of the head and states coassociativity as a lemma (Lemma 10.4),
but its definition of the label (Definition 10.3) is again stated only for a syntactic object $T \in \mathfrak{T}_{\SOz}$ that contains no trace leaves and for its accessible terms.
The definition assigns no label when the head leaf of the contracted subtree already carries a trace label
(Definition 10.3 presupposes that the head leaf is labeled by an element of $\SOz$; since the label set of Definition 10.1 contains no iterated cancellation, the same label is presumably intended),
although \S10.2 of the paper explicitly allows the extraction of terms that contain trace leaves.
The components of all workspaces, contraction quotients included, are assumed to carry head functions (p.~48), but how the head function of $T/_cT_v$, or of an extracted term $T_v$, is related to $h_T$ is not stated.
The proof of the lemma takes the label arising from the second contraction to be the value of the head function of the original tree $T$,
and takes the head function of an extracted term to be the restriction of $h_T$, without stating either as an assumption.

Coassociativity holds when the definition is supplemented by the following reading:
(i) the cancellation symbol is not iterated, so that the same label is assigned from the original lexical item;
(ii) an extracted term carries the restriction of $h_T$, and the contraction quotient $T/_cA$ carries the head function induced by $h_T$,
which sends an internal vertex $u$ to $h_T(u)$ if that leaf survives and otherwise to the trace leaf $a \in A$ below which $h_T(u)$ lies; and
(iii) as for MCB's $\cop_c$ (\S\ref{sec:merge}), trace leaves are not counted among the accessible terms, a restriction that the paper imposes only on the selection of the arguments of Merge ((10.3), p.~49), not on the coproduct.
Under this reading, restriction commutes with the induction and the induction is transitive, so the factors of corresponding terms in Appendix~\ref{sec:proofs-coassoc} agree as trees with head functions,
and Proposition~\ref{lem:label-invariance} below applies, with a tree understood together with its head function and with $\lambda(X)$ the canceled lexical item of the head leaf of $X$.

In what follows we consider the case in which, as in MCB's definition, the label records the subtree,
and examine three possible ways of extending the definition to trees that already contain trace leaves.
Trace leaves themselves are not counted among the accessible terms (they are not extracted).
\begin{description}
\item[Forgetting the labels] The labels of all trace leaves are collapsed to a single symbol $\tr{\star}$. Write $\HAb$ for the resulting space of forests, $\pi\colon\HA\to\HAb$ for the surjection that forgets the labels, and $\copb_c$ for the coproduct on $\HAb$.
  Then $(\pi\ten\pi)\circ\cop_c = \copb_c\circ\pi$ holds for each of the extensions below, and $\copb_c$ is coassociative (Proposition~\ref{lem:label-invariance} below, with $\lambda$ constant).
\item[The literal extension] The subtree to be contracted is used as the label as it stands at that moment (including the trace leaves inside it).
  In this case $\cop_c$ is not coassociative. For $T_w \supset F$, the trace leaf obtained by extracting $F$ and then $T_w/_cF$ is $\tr{T_w/_cF}$,
  whereas the one obtained by extracting $T_w$ at once is $\tr{T_w}$, and the two do not agree.
  The minimal example is the tree $T = \{\alpha,\{\beta,\gamma\}\}$ of Proposition~\ref{prop:delta-d}:
  $(\cop_c \ten \idm)\circ\cop_c(T)$ contains $\beta \ten \{\tr{\beta},\gamma\} \ten \{\alpha, \tr{\{\beta,\gamma\}}\}$,
  whereas $(\idm \ten \cop_c)\circ\cop_c(T)$ contains $\beta \ten \{\tr{\beta},\gamma\} \ten \{\alpha, \tr{\{\tr{\beta},\gamma\}}\}$ instead.
\item[The restoring extension] Every trace leaf $\tr{S}$ inside the subtree to be contracted is restored to $S$ before the subtree is used as the label.
  Writing $\mathrm{res}(X)$ for the tree obtained from a tree $X$ by restoring all the trace leaves in it, the label of the trace leaf that arises from the contraction of $X$ is $\tr{\mathrm{res}(X)}$.
  In the example above, the label for the contraction of $\{\tr{\beta},\gamma\}$ becomes $\tr{\{\beta,\gamma\}}$, and the two terms agree.
\end{description}

\begin{proposition}
\label{lem:label-invariance}
Suppose that trace leaves are not counted among the accessible terms, and that the label of a trace leaf is given by a rule $\lambda$ that depends only on the contracted subtree,
with $\cop_c$ considered on the space spanned by the forests of trees whose leaves are lexical items or trace leaves labeled by values of $\lambda$.
Then $\cop_c$ is coassociative on this space if and only if $\lambda(X/_cF) = \lambda(X)$ holds for every such tree $X$ and every subforest $F$ consisting of accessible terms of $X$.
That is, coassociativity holds exactly when the label does not record what has already been extracted from inside the subtree.
\end{proposition}

The proof is given in Appendix~\ref{sec:proofs-coassoc}. The point is that corresponding terms on the two sides of the coassociativity law agree in their first and second factors whatever $\lambda$ is,
and can differ only in the label of a trace leaf in the third factor.
The three extensions are decided by this criterion. The forgetting of labels ($\lambda$ constant) satisfies the condition. The literal extension does not, as the example above shows.
For the restoring extension, $\mathrm{res}(X/_cF) = \mathrm{res}(X)$ holds, since $\mathrm{res}$ restores each trace leaf created by the contraction of $F$ to $\mathrm{res}$ of the subtree from which it arose,
which is what $\mathrm{res}(X)$ has at that position. Hence:

\begin{proposition}
\label{prop:resolved}
If trace leaves are not counted among the accessible terms and the labels of trace leaves are defined by the restoring extension, then $\cop_c$ is coassociative.
\end{proposition}

Thus the coassociativity of $\cop_c$ that MCB asserts does hold under the restoring extension. In what follows, $\cop_c$ is understood with this extension.
Under it, the label of a trace leaf is always a syntactic object without trace leaves. Accordingly, we enlarge $\HA$ to the space spanned by the forests of trees whose leaves are either lexical items or trace leaves carrying such labels; this space is closed under $\cop_c$.
The same criterion applies to the head-label rule of \S\ref{sec:ms}, which satisfies the condition under the reading supplied above.

The problem that remains for $\cop_c$ is connectedness (the situation is the same whether the labels are retained or forgotten).
The grading by the number of vertices, which makes the Connes--Kreimer Hopf algebra of rooted trees connected and works for $\cop_\rho$, is not available for $\cop_c$:
contraction keeps the root $v_i$ of each $T_{v_i}$ as a leaf, so $\#V(T/_cF_v) = \#V(T) - \#V(F_v) + j$ for a subforest with $j$ components, and the number of vertices is not preserved.
MCB therefore gives two other gradings.
\begin{itemize}
\item In Chapter 1 the grading is by the number of edges, and $(\HA, \sqcup, \cop_c)$ is said to be a graded bialgebra \mcb{Lemma 1.2.10, p.~37}. But trees without edges (trees consisting of a single lexical item $\alpha \in \SOz$) fall into degree $0$, so the bialgebra is not connected.
  MCB acknowledges this and asserts that taking the quotient by the ideal generated by $1 - \alpha$ ($\alpha \in \SOz$) makes degree $0$ one-dimensional and yields a Hopf algebra
  \mcb{pp.~37--38}. This quotient, however, does not work:
  since $\cou(1-\alpha) = 1 \neq 0$, the ideal $\langle 1-\alpha \rangle$ is not a coideal, and the counit does not descend to the quotient.
  MCB calls $\alpha$ a group-like element, but under \eqref{eq:coproduct} we have $\cop_c(\alpha) = \alpha \ten 1 + 1 \ten \alpha$, so $\alpha$ is primitive;
  the quotient would work only if the single-vertex tree $T/_cT$ were identified with $1$, an identification that MCB does not make.
\item In Chapter 3 the grading is by the number of leaves, and the structure is declared to be
  ``\textit{a commutative, associative, coassociative, graded, and connected Hopf algebra}'' \mcb{\S3.1.4, p.~210}.
  But if trace leaves are counted, the grading is incompatible with $\cop_c$ (the $k$ leaves of $T_v$ and the $n-k+1$ leaves of $T/_cT_v$ add up to $n+1 \neq n$);
  if they are not counted, the grading is compatible, but trees consisting of trace leaves only (for instance $\{\tr{\alpha},\tr{\beta}\}$) fall into degree $0$, so the bialgebra is not connected.
\end{itemize}
Thus neither of the gradings that MCB gives yields connectedness, and a structure that simultaneously satisfies the five properties demanded in Chapter 3
(commutative, associative, coassociative, graded, and connected) is not constructed within MCB.

The connectedness obstacle can be removed by revising the grading as proposed below.
The possible gradings are not limited to the two adopted by MCB.
The number of edges and the number of non-trace leaves are both gradings compatible with $\cop_c$, and so is their sum.
With the sum, the only basis elements left in degree $0$ are the forests all of whose components are isolated trace leaves (including the empty forest).

\begin{proposition}
\label{prop:grading}
For $\cop_c$ with the extension of Proposition~\ref{prop:resolved}, define the degree of a forest $F$ by
\[
\deg(F) \defequiv \#E(F) + \#\{\text{non-trace leaves of $F$}\},
\]
where $E(F)$ is the set of edges of $F$.
The subspace $\HA^{\circ} \subset \HA$ spanned by the forests that have no isolated trace leaf as a component is closed under $\sqcup$ and $\cop_c$,
and $(\HA^{\circ}, \sqcup, \cop_c, \deg)$ is a commutative graded connected bialgebra. Hence it is a Hopf algebra.
\end{proposition}
The proof is given in Appendix~\ref{sec:proofs-grading}.
The connectedness obstacle on the $\cop_c$ side is therefore a flaw in MCB's particular choice of grading, not a defect of the algebraic structure itself.
Thus $\cop_c$, once a rule for the labels of trace leaves and a grading are supplied, becomes a graded connected Hopf algebra while each trace leaf retains the structure of the contracted subtree.
What MCB's argument lacked was these two additional specifications.

\subsection{Interim Summary}
\label{sec:coproducts-summary}
\begin{table}[t]
\centering
\small
\renewcommand{\arraystretch}{1.15}
\begin{tabular}{@{}p{0.11\textwidth}p{0.19\textwidth}p{0.20\textwidth}p{0.15\textwidth}p{0.24\textwidth}@{}}
\toprule
Coproduct & Right channel & Coassociativity & Connected grading & Remarks \\
\midrule
$\cop_\rho$ & a vertex remains at each cut (not binary) & holds & number of vertices & the Connes--Kreimer coproduct itself; MCB gives a proof \\
$\cop_d$ & deleted and reduced to a binary tree & fails (Proposition~\ref{prop:delta-d}) & --- & acknowledged by MCB; the separate paper has not appeared; Drinfeld's quasi-Hopf algebras are unavailable (Proposition~\ref{prop:quasi-hopf}) \\
$\cop_c^{h}$ & a trace leaf remains (head label) & holds \citep{MarcolliSkigin2025}, with the reading of \S\ref{sec:delta-c} & connected once revised (Proposition~\ref{prop:grading}) & rule given in a paper following MCB; its definition needs supplementing \\
Xiao's coproduct & deleted and reduced to a binary tree (cuts restricted) & holds \citep{Xiao2026} & number of leaves & comes with a proof from the outset \\
$\cop_c$ & a trace leaf remains & holds once a rule is supplied (Proposition~\ref{prop:resolved}) & connected once revised (Proposition~\ref{prop:grading}) & MCB gives neither a rule nor a proof; fails under the literal extension \\
\bottomrule
\end{tabular}
\caption{The five coproducts and the axioms of a Hopf algebra.}
\label{tab:four-coproducts}
\end{table}
In this section we have proposed, for $\cop_c$, the additional assumptions that are needed for it to satisfy the conditions on the coproduct of a Hopf algebra.
As a result, four of the five coproducts taken up in this section yield Hopf algebras
(for $\cop_\rho$, however, the Hopf algebra is not $\HA$ but a larger space that also contains forests of non-binary trees).

\section{Five Coproducts and Traces}
\label{sec:binding}

The two properties of the previous section, coassociativity and connectedness, can be satisfied by four of the five coproducts, all but $\cop_d$, once a rule and a grading are supplied,
and in that respect these four cannot be ranked against one another.
As we saw in \S\ref{sec:two-coproducts} and \S\ref{sec:not-hopf}, however, what distinguishes the five coproducts is the kind of representation that movement leaves in the remaining tree.
This representation is what the syntactic component hands over to the semantic component.
Since Chomsky himself takes interpretation at the conceptual-intentional interface to be part of Minimalist theory (\S\ref{sec:two-coproducts}),
a description of Minimalism as a Hopf algebra must meet the requirements that semantics has placed on syntactic structure.

Those requirements have not stayed fixed: the relation between semantics and the syntax of Minimalism has changed over time.
In this section we therefore review how moved elements in particular have been treated in syntactic structure, and make explicit the requirements that bear on them.
We then show that none of the five coproducts of the previous section meets them.

\subsection{From Traces and Indices to Copies}
\label{sec:copies}
In generative grammar, movement (Move-$\alpha$) was originally an operation distinct from Merge. Whether overt or covert, a moved element is assigned an index and adjoined to a structure containing it, and a trace bearing the same index remains in its original position.
This is the data structure of the output of syntax that standard formal semantics, such as \citet{HK1998} and \citet{Buring2005}, presupposes.
The moved element introduces a binder $\lambda_i/\mu_i$ bearing the same index $i$ as the trace $t_i$, and the binder and the trace $t_i$ are interpreted in the semantic component as $\lambda x_i$ and $x_i$, respectively.
Binding theory rests on this correspondence.

\citet{Chomsky1995} adopted the copy theory of movement, on which what is left in the original position is a copy of the moved element (ch.~3, p.~202; first published in 1993).
The motivation was to eliminate reconstruction as an operation: the copy remains at LF, ``\textit{providing the materials for `reconstruction.'}''
The same work went on to propose the Inclusiveness Condition, according to which syntactic computation must not introduce any element beyond what lexical items already contain (p.~228).
Indices violate this condition.
Chomsky, however, remarked in the same chapter that marking as distinct the syntactic objects formed by separate selections of the same lexical item is a departure from the Inclusiveness Condition, but
``\textit{one that seems indispensable}'' (p.~227).
Movement was later reconceived as one case of Merge (Internal Merge):
Merge does not alter the original structure (the No Tampering Condition), and an element merged with a structure containing it appears in two positions.

\subsection{Semantic Interpretation under the Copy Theory}
\label{sec:copy-interp}
Whether formal semantics can be maintained under this index-free theory became an issue early on.
\citet{Fox2002} proposed Trace Conversion as an interpretive operation for the copy theory. The moved element is interpreted as it is in the upper position,
and the following two operations apply to the lower copy \citep[p.~67]{Fox2002}.
\begin{itemize}
\item Variable Insertion: the predicate $\lambda y(y = x)$, ``being identical to the variable $x$,'' is added to the predicate of the lower copy.
\item Determiner Replacement: the determiner of the lower copy is replaced by \textit{the}.
\end{itemize}
For example, when \textit{everyone} raises, the lower copy becomes the definite description ``the person identical to $x$,'' and $\lambda$-abstraction over $x$ takes place in the upper position.
This operation demands two things of the output of syntax.
One is that the lower copy retain its predicate (\textit{person} in the example above).
The other is the identity of the variable $x$. Fox writes $x$ without an index, and does not say how the variable inserted into the lower copy is matched with the one bound by the $\lambda$-abstraction above.
Yet this $x$ must be the same variable as the one abstracted over under the upper element,
and when more than one element has moved, their variables must be kept distinct from one another.
That is, the information as to which lower copy has the same variable as which upper element must be readable from the output of syntax.
Trace Conversion, therefore, does not in fact go through without something equivalent to an index.

Chomsky, for his part, states that, when two elements of the same form occur, one must distinguish whether they are two copies of a single element or separate elements (repetitions).
The two occurrences of \textit{what} in \textit{what hit what?} are repetitions, whereas the two in \textit{what was hit what?} are copies \citep[p.~40]{Chomsky2013}.
As for the means of drawing the distinction, Chomsky and his colleagues suggest that one need only see, within a phase, whether the element in question was formed by Internal Merge,
while noting that this approach ``\textit{remains to be formalized}'' \citep[p.~247]{ChomskyGallegoOtt2019}.
Later Chomsky posits FormCopy as an operation that establishes the copy relation between structurally identical elements \citep[pp.~17, 20]{Chomsky2021GK}.
\citet{CollinsGroat2018} review the proposals for distinguishing copies from repetitions (added structure such as indices, multidominance, and chains; configurational approaches; and phase-level memory)
and conclude that ``\textit{every proposal we examine faces serious shortcomings}'': indices and multidominance violate the Inclusiveness Condition,
and none of the proposals explains the distinction, ``\textit{but merely provide}[s] \textit{stipulations capturing it}.''
Moreover, what these proposals address goes no further than the distinction between copies and repetitions.
How the semantic interpretation of Quantifier Raising and of wh-movement in relative clauses---that is, the interpretation with a binder above and a variable below---is to be obtained without indices is not discussed there.

Chomsky himself was aware of this problem. As noted above, he called the marking that distinguishes occurrences of the same lexical item something that seems indispensable \citep{Chomsky1995},
and later stated that there must be some way to identify an internally merged element with its copy, but not with other items that have the same feature composition, a problem he took to be solved by phase-level memory (Chomsky 2008, as quoted in (36) of \citealp{CollinsGroat2018}).
More recently he has called the distinction between repetitions and copies ``\textit{a long-standing and problematic issue that has been left in some obscurity}'' (\citealp[p.~26]{Chomsky2021GK}; \citealp[p.~24]{Chomsky2024GE}).
The latter also states what must hold: ``\textit{At CI, all copies must have precisely the same interpretation}.''
In \textit{The man arrived}, the subject \textit{the man} and its unpronounced lower copy after the verb must refer to the same man,
whereas the two occurrences of \textit{the man} in \textit{The man met the man} are ``\textit{structurally identical elements that are not copies}'' \citep[p.~25]{Chomsky2024GE}.
In an earlier paper, Chomsky and his colleagues write of the chain of occurrences created by wh-movement that it is ``\textit{interpreted as an operator-variable dependency},''
and they spell the interpretation out with variables: ``\textit{which book x about y is such that this law y has triggered the publication of x}'' \citep[p.~235]{ChomskyGallegoOtt2019}.
Where these variables come from in the output of syntax, however, is not stated.
In the Minimalism of this period, partly because the framework itself was repeatedly revised and its specification never settled, it sufficed to state one's view on such points informally.
In recent proposals, successive-cyclic movement in the traditional sense is itself being reconsidered \citep[p.~32]{Chomsky2023MiracleCreed}, and it is unclear how Quantifier Raising is to be realized in that framework.

\subsection{The Coproduct Formalization Leaves No Such Room}
MCB defines algebraically, by means of a coproduct, what becomes of a tree as the result of movement.
Whichever coproduct is chosen, what remains in the lower position is fixed in one form. As a result, there is no longer any room for stating one's view informally.
This makes it difficult to continue Chomsky's long-standing practice of leaving binding theory to the semantic side.
We summarize what semantic interpretation demands of the representation that results from movement in the following two conditions, both stated from the standpoint of the lower position.
\begin{description}
\item[Retention] The structure that the lower position had before movement can be recovered from the representation.
\item[Identifiability] In interpreting the lower position, it can be determined from the representation which upper element, if any, corresponds to it.
\end{description}
Movement with indices satisfies both conditions. In \citet{HK1998} and \citet{Buring2005}, an indexed trace $t_i$ stands in the lower position, and the moved element stands above together with a binder bearing the same index.
The structure that the lower position had before movement is recovered by looking above for the constituent with the same index, so Retention is satisfied.
The trace $t_i$ is interpreted as the variable $x_i$ and the binder as abstraction over $x_i$; the two are related through the assignment function, so Identifiability is satisfied as well.

The copy theory satisfies Retention only. A copy remains in the lower position, so the structure is there without having to be recovered.
The correspondence with an upper element, however, goes no further than identity of structure: when several constituents have the same structure, it is not determined which of them, if any, the lower copy corresponds to.
This is the problem of copies and repetitions (\S\ref{sec:copy-interp}).
The Trace Conversion of \citet{Fox2002} does not remove this problem; it presupposes Identifiability.
Trace Conversion is stated as an operation on chains: Fox writes that ``\textit{the copy at the tail of the chain can be converted to an element that contains a variable in various ways}'' \citep[p.~67]{Fox2002}, and its two operations apply to this copy.
Which constituent is the tail of which chain, and hence which $\lambda$ above binds the variable to be inserted, is thus taken as given in the input.
Fox does not say where in the representation this information is to be found: as we saw in \S\ref{sec:copy-interp}, he writes the variable without an index and does not state how it is matched with the $\lambda$-abstraction above.
What Trace Conversion shows, therefore, is how a representation that satisfies both conditions is interpreted under the copy theory, not how Identifiability is obtained.

In terms of these two conditions, the copy theory (\S\ref{sec:copies}) secures Retention, and it is the Inclusiveness Condition that leaves no means of securing Identifiability:
the copy remains in the lower position with its structure, but the indices that would link it to the upper element have no place.
To show that the correct interpretation is not obtained when Retention or Identifiability fails, we give three elementary examples.

\begin{example}[Relative clauses]
\label{exa:relative}
In \textit{girl who John loves}, the relative clause \textit{who John loves} denotes the one-place predicate $\lambda x.\,\mi{love}(j,x)$, ``the $x$ that John loves,'' and is conjoined (or, equivalently, intersected) with the predicate denoted by \textit{girl}.
This predicate is obtained by turning the lower position into a variable and abstracting over it in the upper position: as \citet[p.~75]{Fox2002} puts it, movement turns the relative clause into a predicate.
If there is no element in the lower position, there is nothing to turn into a variable; and merely interpreting \textit{who} in the lower position yields no more than \textit{John loves who}. Either way, no one-place predicate is obtained.
MCB's rule $s(\tr{T_v}) \defequiv s(T_v)$ \mcb{\S3.9.1, pp.~328--329}, which gives the trace leaf the value of the moved element, amounts to this ``interpretation in the lower position.''
The variable in the lower position must be chosen to be the very one abstracted over above; it is Identifiability that settles this choice.
\end{example}

\begin{example}[Inverse scope of quantifiers]
\label{exa:qr}
Consider \textit{A girl who hates every boy kissed every boy}. In \citet{Buring2005}, the subject is interpreted as it is in its surface position,
and a quantifier can be interpreted in situ by type shifting (the operator $\kappa$); the occurrence of \textit{every boy} inside the relative clause is interpreted in this way.
For the reading in which the object \textit{every boy} takes scope over the subject (inverse scope), however, Quantifier Raising is required.
The logical form is then
\[
[\,\text{every boy}\ [\,\mu_1\ [\,[\,\text{a girl who hates every boy}\,]\ [\,\text{kissed}\ t_1\,]]]]
\]
and the meaning obtained is
\[
\forall y\,\exists x\,[\,\mi{girl}(x) \wedge \forall z\,\mi{hate}(x,z) \wedge \mi{kiss}(x,y)\,],
\]
where $y$ and $z$ range over boys: for every boy there is a girl who hates every boy and kissed him.
Now suppose that the indices are removed and a copy remains in the lower position. There are then two occurrences of \textit{every boy} below the upper \textit{every boy}: the lower copy of the object, and the one inside the relative clause, which in the system of \citet{Buring2005} has not moved and corresponds to no upper element.
Which of the two corresponds to the upper \textit{every boy} cannot be determined from the form of the words.
If the one inside the relative clause is taken to correspond to it, and the object is interpreted in situ, what is obtained is
\[
\forall y\,\exists x\,[\,\mi{girl}(x) \wedge \mi{hate}(x,y) \wedge \forall z\,\mi{kiss}(x,z)\,],
\]
that is, for every boy there is a girl who hates him and kissed every boy, which is a different truth condition.
When Identifiability fails, the meaning is not determined.
\end{example}

\begin{example}[Condition C]
\label{exa:condC}
Consider the following contrast \citep[pp.~65, 68]{Fox2002}.
\begin{enumerate}
\item[(i)] ??/*\textit{Someone introduced him$_1$ to every friend of John's$_1$.}
\item[(ii)] \textit{Someone introduced John$_1$ to every friend of his$_1$.}
\end{enumerate}
In (i), Quantifier Raising can move the object \textit{every friend of John's} to a position higher than \textit{him}, as the availability of inverse scope in (ii) shows.
In the landing position \textit{him} does not c-command \textit{John}, so that if only the structure after raising were inspected, with the lower position a variable without internal structure, (i) would wrongly be predicted not to violate Condition C.
Under Trace Conversion, the lower copy becomes ``the friend of John's identical to $x$'' and thus contains \textit{John}; hence \textit{him} c-commands \textit{John}, and the Condition C violation is correctly predicted.
In frameworks that distinguish the Spell-Out structure from LF, the same result is obtained by checking c-command before Quantifier Raising applies.
Either way, the structure that the lower position had before movement must be recoverable. This is what Retention requires.
\end{example}

\subsection{Evaluation of the Five Coproducts}
From this standpoint, we reexamine the five coproducts of the previous section.
\begin{description}
\item[$\cop_d$ and Xiao's coproduct] In both, the right channel is the deletion quotient, and nothing is left in the lower position. Since the lower position itself is absent from the representation, neither Retention nor Identifiability is satisfied, and neither the method of \citet{HK1998} nor that of \citet{Fox2002} can be applied.
  The terms of Xiao's coproduct correspond one-to-one to subsets of the leaves, and where the extracted part was located in the original tree is not recorded in the right factor.
  As a solution to the coassociativity problem of $\cop_d$, this is algebraically the cleanest one, but it is the farthest from the two conditions.
\item[$\cop_\rho$] The upper vertex of each cut edge remains, so the lower position is marked. That vertex, however, has neither internal structure nor a label:
  there is no clue either for recovering the structure it had or for determining which upper element it corresponds to. Neither Retention nor Identifiability is satisfied.
\item[$\cop_c^{h}$] A trace leaf is left in the lower position, but its label is only the lexical item of the head of the contracted subtree; from the label alone, no more than the head can be recovered.
  \citet{MarcolliSkigin2025} hold that the label need not retain the structure, since the extracted subtree is still present in the left factor of the coproduct (see the quotation in \S\ref{sec:ms}).
  The subtree is indeed there. For its structure to be recovered for the lower position, however, as the determination of Condition C in Example~\ref{exa:condC} requires, it must be determined which element the trace leaf corresponds to;
  and since the trace leaves of moved elements with the same head receive the same label, this cannot be done. Neither Retention nor Identifiability is satisfied.
\item[$\cop_c$] A trace leaf (what MCB calls a trace) is left in the lower position, and its label records the structure of the contracted subtree.
  In the version that forgets the labels of trace leaves ($\copb_c$), no structure is recorded, and the situation is that of $\cop_\rho$.
  Under the restoring extension (Proposition~\ref{prop:resolved}), the structure that the lower position had before movement can be recovered from the label; to that extent Retention is satisfied.
  As \citet{Collins2026} points out, however, the structure is only recorded in the label; the lower position itself is a single leaf.
  For the conditions of binding theory to apply, they must be redefined so as to look into the structure of the subtree recorded as the label.
  Even then, since no element of the tree bears an index, looking into the label does not reveal which elements outside it the elements inside it are related to.
  Furthermore, the label of a trace leaf records structure, not an index.
  Every element elsewhere in the tree that has an isomorphic structure is a candidate for the element corresponding to that trace leaf. In Example~\ref{exa:qr}, there are two occurrences of \textit{every boy} besides the trace leaf $\tr{\text{every boy}}$, the raised one and the one inside the relative clause, and which of them corresponds to it is not determined.
  Hence Identifiability is not satisfied.
\end{description}
To sum up, Retention is satisfied only by $\cop_c$, and there only in the form of a label; Identifiability is satisfied by none of the five.
Neither $\cop_d$, which is not coassociative, nor any of the four coassociative coproducts satisfies both conditions.
Together with coassociativity (\S\ref{sec:not-hopf}), the results are summarized in Table~\ref{tab:four-coproducts-traces}.

\begin{table}[t]
\centering
\small
\begin{tabular}{@{}llll@{}}
\toprule
Coproduct & Coassociativity & Retention & Identifiability \\
\midrule
$\cop_\rho$ & holds & no (position only) & no \\
$\cop_d$ & fails & no (nothing remains) & no \\
$\cop_c^{h}$ (head labels) & holds & no (head only) & no \\
Xiao's coproduct & holds & no (nothing remains) & no \\
$\cop_c$ (restoring extension) & holds & only as a label & no \\
\bottomrule
\end{tabular}
\caption{The five coproducts: coassociativity, and the two conditions, Retention and Identifiability, that semantic interpretation imposes on the representation that results from movement.}
\label{tab:four-coproducts-traces}
\end{table}

\section{Birkhoff Factorization Revisited}
\label{sec:birkhoff}

MCB claims that the Birkhoff factorization is the mechanism of the syntax--semantics interface, and this forms the other pillar of MCB's claim that ``language and physics obey the same laws.''
A character $\phi : \HA \to \RBalg$ ($\RBalg$ a Rota--Baxter algebra or a semiring) is factored, by the Bogolyubov recursion,
as $\phi = \phi_-^{\conv(-1)} \conv \phi_+$, or equivalently $\phi_+ = \phi_- \conv \phi$ (the form that survives in the semiring version, where $\phi_-$ has no convolution inverse),
where $\phi_+$ carries ``the consistent part'' and $\phi_-$ ``the part to be removed.''
The theorems MCB invokes are the algebra version due to Connes--Kreimer \mcb{Prop.~3.1.7, p.~213} and the semiring version due to Marcolli--Tedeschi \mcb{Prop.~3.1.9, pp.~213--214},
both of which presuppose a commutative graded connected Hopf algebra.
As we saw in \S\ref{sec:not-hopf}, this presupposition is not met in the setting of MCB's own text:
$\cop_d$ is not coassociative, and $\cop_c$ lacks a rule for the labels of trace leaves and is not connected under MCB's gradings
(MCB's Chapter~3 works with $\cop_c$, not with $\cop_\rho$).

MCB realizes the Birkhoff factorization by a different route.
The Bogolyubov recursion evaluates $\phi_-$ only on the left factors of the reduced coproduct, and the left factors are proper subforests,
so the recursion terminates by induction on the number of vertices even without coassociativity or connectedness.
MCB itself states: ``\textit{we will focus only on the bialgebra structure and not discuss the antipode map explicitly}'' \mcb{p.~159}, and
``\textit{the factorization that we consider does not explicitly use the antipode}'' \mcb{p.~365}.
The multiplicativity of $\phi_\pm$, the substance of the factorization, also follows from the Rota--Baxter identity and the multiplicativity of $\cop$ alone
(Lemma~\ref{lem:multiplicative} in Appendix~\ref{sec:proofs-bogolyubov}; the semiring version holds likewise under the Rota--Baxter identity of weight $+1$: Lemma~\ref{lem:multiplicative-semiring}).
That is, MCB's ``theoretical construction of the same form as in quantum field theory'' stands not as a consequence of the Hopf-algebra theorem,
but as a consequence of a weaker statement that requires neither coassociativity nor connectedness (Lemma~\ref{lem:multiplicative}).

Which results use coassociativity can be checked mechanically (Table~\ref{tab:coassoc-use}).
The associativity of the convolution product is coassociativity itself, and the group of characters and the uniqueness of the antipode follow from it.
The existence of the antipode by Takeuchi's theorem cannot even be formulated without coassociativity, since the iterated coproduct $\cop^{(n)}$ appears in the formula.
On the other hand, the existence of the Bogolyubov recursion and the multiplicativity of $\phi_\pm$ do not use coassociativity.
In other words, what survives the loss of coassociativity is only the part that MCB already uses as a direct construction,
and what is lost is the part that makes a Hopf algebra a Hopf algebra (the group of characters and the antipode).

\begin{table}[t]
\centering
\small
\begin{tabular}{@{}p{0.50\textwidth}p{0.42\textwidth}@{}}
\toprule
Result & Uses coassociativity? \\
\midrule
Associativity and unit of the convolution product; group of characters & Yes (it is associativity itself) \\
Uniqueness of the antipode & Yes (via the associativity of convolution) \\
Takeuchi: graded connected $\Rightarrow$ existence of the antipode & Yes (the iterated coproduct $\cop^{(n)}$ appears in the formula) \\
Existence of the Bogolyubov recursion & No (induction on the number of vertices of the left factor) \\
Multiplicativity of $\phi_\pm$ & No (Rota--Baxter identity and multiplicativity of $\cop$) \\
\bottomrule
\end{tabular}
\caption{Main results on Hopf algebras and their dependence on coassociativity.}
\label{tab:coassoc-use}
\end{table}

The factorization itself is obtained, but the theorems MCB invokes are not at work there. With respect to the analogy with physics, this is a step backward.
In contrast, $\cop_c$ supplemented with the restoring extension and the grading of Proposition~\ref{prop:grading} (on $\HA^\circ$; Propositions~\ref{prop:resolved} and \ref{prop:grading}) yields a commutative graded connected Hopf algebra,
and so, according to \citet{Xiao2026}, does the coproduct defined in that paper.
The premises of the two theorems MCB invokes are therefore satisfied literally.
For characters with values in a Rota--Baxter algebra the theorem of Connes--Kreimer \mcb{Prop.~3.1.7, p.~213} applies,
and for characters with values in a Rota--Baxter semiring the theorem of Marcolli--Tedeschi \mcb{Prop.~3.1.9, pp.~213--214} applies, each as a theorem.
In the algebra-valued case, moreover, the fact that characters form a group under the convolution product, and inverses given by the antipode, become available.
What changes here is not whether the factorization exists. As we saw above, the existence of the semiring-valued factorization requires neither coassociativity nor connectedness.
What changes is whether that factorization is an instance of a general theorem.

\section{Conclusion and Outlook}
\label{sec:direction}

This paper started from the question whether Merge forms a Hopf algebra. We summarize the answers obtained here for the five coproducts that appear in the literature.
First, the coassociativity of the coproduct is not an essential obstacle: $\cop_\rho$ and the coproduct of \citet{Xiao2026} have been shown to be coassociative,
and the coproduct $\cop_c^{h}$ of \citet{MarcolliSkigin2025} is coassociative once its definition of the labels is supplemented by the reading given in \S\ref{sec:delta-c}.
$\cop_c$, too, becomes a graded connected Hopf algebra once the rule for the labels of trace leaves and the grading are supplied as proposed in this paper (Propositions~\ref{prop:resolved} and \ref{prop:grading}).
The only one that is not coassociative is $\cop_d$, which MCB uses in the definition of Merge (\S\ref{sec:not-hopf}).
Whether Merge \eqref{eq:merge}, rewritten with one of the coassociative coproducts, still agrees with the Merge of Chomsky and his colleagues is, however, a separate question;
for $\cop_c$ and $\cop_\rho$ the newly formed syntactic object is the same and only the shape of the remaining workspace differs (\S\ref{sec:two-coproducts}).

On the other hand, the standard semantic interpretation in terms of variables and binding demands two things of the representation that results from movement: Retention, that the structure the lower position had before movement be recoverable, and Identifiability, that, in interpreting the lower position, it be determinable which upper element, if any, corresponds to it.
We have shown in this paper that none of the five coproducts meets both conditions (\S\ref{sec:binding}, Table~\ref{tab:four-coproducts-traces}).

As argued in this paper, Identifiability is not a property that was lost for the first time in MCB; it is a property that Minimalism has tacitly lacked ever since the Inclusiveness Condition of \citet{Chomsky1995} left no place for indices in the copy theory.
Retention, by contrast, is a property that the copy theory did satisfy; of the five coproducts examined here, only $\cop_c$ satisfies it, and only in the form of a label.
In this respect these coproducts, as they stand, neither amount to a formalization of Minimalism nor make up for what Minimalism itself lacks.
What MCB has brought about is that the problem is now posed in a formal shape.
Whether the working hypothesis H holds also depends on whether the ``syntax of Merge'' that it speaks of is taken to include the representations that semantic interpretation presupposes, namely representations satisfying Retention and Identifiability.

There are at least three directions in which to proceed from here.
\begin{enumerate}
\item[(A)] Not to require Retention and Identifiability of syntactic structure. Priority is given to the Hopf algebra and to the analogy with physics that it brings, and the connection with the standard semantic interpretation in terms of variables and binding is given up.
  The theory of meaning is then confined to semantics that needs neither. One candidate is distributional semantics, which takes the semantic space to be a linear space;
  that MCB takes its semantic spaces to be vector spaces \mcb{\S3.2.3, p.~234; \S3.10, p.~338} fits this direction well.
\item[(B)] To design a new coproduct that satisfies Retention and Identifiability. Such a coproduct can be constructed\footnotemark: if, in $\cop_c$, every vertex is decorated with an index, distinct for distinct vertices of the original tree, and each trace leaf carries the index of the contracted vertex,
  one obtains a coproduct that is coassociative, forms a graded connected Hopf algebra, and satisfies both conditions. The proofs, and the application to Merge, Minimal Yield, and semantic interpretation, are left for another occasion.
  This direction, however, gives up the Inclusiveness Condition. That contraction violates the No Tampering Condition is inherited from MCB's $\cop_c$.
\item[(C)] Not to require a Hopf algebra. The analogy with physics is given up, and a separate algebraic formulation is set up that conforms to the specification of Chomsky's Merge, in which both the upper and the lower copy are retained.
  The form of MCB's Merge survives on this route: the action of Merge~(\ref{eq:merge}) applies the coproduct only once, and coassociativity plays no role there;
  and the Birkhoff factorization at the syntax--semantics interface requires neither coassociativity nor connectedness (\S\ref{sec:birkhoff}).
\end{enumerate}
\footnotetext{%
As for direction (B), \citet{Andrews2026} proposes a different construction, one that keeps the lower copy: the extracted subtree is not contracted but left in place,
and its root is recorded, as part of the workspace, in a set of canceled vertices; the correspondence between the upper and the lower copy
is determined by assigning a unique index to each lexical item, following \citet{CollinsStabler2016}.
Andrews argues that the resulting coproduct is coassociative and forms a graded connected Hopf algebra, though one whose graded components are not finite-dimensional.
}%
These three are not exhaustive, but further routes are not obvious either. For instance, one could keep the correspondence between the upper and the lower element outside the elements of the algebra, as part of the history of the derivation.
Chomsky, however, rules this out on the grounds that derivations are Markovian and have no access to earlier stages:
``\textit{If derivations were non-Markovian, with access to earlier stages, then the rule of copy formation CF could be restricted to IM-configurations $\langle X,Y\rangle$. But the Markovian property bars that easy answer}'' \citep[p.~24]{Chomsky2024GE}.

Whichever direction is taken, describing the action of Merge as the composite of coproduct, projection, grafting, and product has separated the four stages of the enumeration of Merge candidates, the selection of arguments, so-called ``set formation,'' and the return to the workspace (\S\ref{sec:merge}),
and has made it possible to discuss the difference between the translations of movement as a difference between coproducts (\S\ref{sec:two-coproducts}).
How the correspondence between the upper and the lower copy is to be determined without indices has, since the adoption of the copy theory, been left to semantics, with no specification settled on the side of syntax.
What could be dealt with, in an informal description, by writing that a chain is ``interpreted as an operator-variable dependency'' must, as soon as it is written with a coproduct, answer the question of where that variable comes from.
This is a criticism of MCB, and at the same time a new line of discussion that MCB's formalization has brought about.

\section*{Changes from the First Version}
The first version of this paper (arXiv:2611.11502v1) contained the following errors.
First, it stated that there are ``as far as we know'' four ways to recover a Hopf algebra, but it did not include the constructions of \citet{MarcolliSkigin2025} and \citet{Xiao2026}.
Second, it stated that $\cop_c$ with the labels retained is not coassociative, but that argument was limited to the case in which the labels of trace leaves are extended literally; coassociativity holds exactly for those rules of extension under which the label does not depend on what has already been extracted (Proposition~\ref{lem:label-invariance}), and the restoring extension is one of them.
Third, it stated that the specification of Minimalism in the working hypothesis H should include the distinction of traces by indices, but Chomsky's Minimalism under the copy theory has no indices, and no traces as separate indexed elements: where the word ``trace'' is still used, it names one kind of copy (see \S\ref{sec:two-coproducts}).
The present version has been reorganized so as to correct these points. %
The examination of the five coproducts in light of what semantic interpretation demands of the representation that results from movement (\S\ref{sec:binding}) is new in this version.

\section*{Acknowledgments}
I am grateful to Naoki Fukui for sharing his view on the current formulation of Minimalism and for bringing \citet{Collins2026} to my attention.
The errors in the first version of this paper could be corrected thanks to these comments.
I am grateful to Kenta Mizutani for informing me of the history of research on Trace Conversion and of the Condition C data showing that the structure of the moved element must be available for the lower position.
I also thank the presenters at the workshop ``Computational Generative Linguistics: Overview and Prospect'' (Linguistic Society of Japan, 15 November 2026),
Yohei Oseki, who also organized and chaired it, Shohei Hidaka, Akihiro Maeda, Yushi Sugimoto, and Sarasa Takeda, for discussion before and at the workshop.
I am particularly grateful to Yushi Sugimoto for sharing the most recent literature with me.
I thank Avery Andrews, Christopher Collins, and Yutaro Ueno for their comments on the first version of this paper, and Andrews and Collins also for sending me their related work.
All remaining errors are my own.
This work was partially supported by JST CREST Grant Number JPMJCR2565, Japan, and JSPS KAKENHI Grant Number JP23H03452, Japan.

\appendix
\section{Proofs}
\label{sec:proofs}

\subsection{Propositions \ref{lem:label-invariance} and \ref{prop:resolved} (coassociativity of the contraction coproduct with labels)}
\label{sec:proofs-coassoc}

\paragraph{Notation.}
Fix a tree $T$ and write $u \leq v$ when the vertex $u$ lies in the subtree $T_v$. A \emph{subforest} of $T$ is either the set consisting of the root alone,
or an antichain $A$ (possibly empty) of vertices of $T$ that are neither the root nor trace leaves. We write $F_A \defequiv \bigsqcup_{a \in A} T_a$ for the corresponding forest.
$T/_cA$ is obtained by contracting each $T_a$ ($a \in A$) to its root $a$, which becomes a trace leaf, and we set $T/_c\{\text{root}\} \defequiv 1$ (the empty forest).
The coproduct is the sum over all subforests, taken as sets of vertices:
\[
\cop_c(T) = \sum_{A} F_A \ten T/_cA
\]
(the term with $A = \{\text{root}\}$ is $T \ten 1$, and the term with $A = \emptyset$ is $1 \ten T$).
Distinct antichains that yield isomorphic forests count as distinct terms; for instance, $\cop_c(\{\alpha,\alpha\})$ contains $\alpha \ten \{\tr{\alpha},\alpha\}$ with coefficient $2$.
The vertices of $T/_cB$ are identified with the vertices of $T$ that do not lie strictly below any element of $B$; among them, the trace leaves are the elements of $B$ and the vertices that were already trace leaves in $T$.
Let $L$ be a set of labels, consider the trees whose leaves are lexical items or trace leaves labeled by elements of $L$, and let $\lambda$ assign an element of $L$ to each such tree up to isomorphism;
$\cop_c$ is defined on the space spanned by the forests of such trees, and the label of the trace leaf at $a$ is $\lambda(T_a)$.
For forests the coproduct is defined multiplicatively, so it suffices to consider a single tree $T$. We assume $\SOz \neq \emptyset$.

\paragraph{Correspondence of terms.}
The terms of $(\cop_c \ten \idm)\circ\cop_c(T)$ are indexed by the pairs $(A,B)$ in which $A$ is a subforest of $T$ and $B$ is obtained by choosing, for each $a \in A$, a subforest of $T_a$;
equivalently, $B$ is a subforest of $T$ each element of which satisfies $b \leq a$ for some $a \in A$. The term is $F_B \ten F_A/_cB \ten T/_cA$.
The terms of $(\idm \ten \cop_c)\circ\cop_c(T)$ are indexed by the pairs $(B,C)$ in which $B$ is a subforest of $T$ and $C$ is a subforest of $T/_cB$. The term is $F_B \ten F_C \ten (T/_cB)/_cC$, with $F_C$ formed in $T/_cB$.
When $B = \{\text{root}\}$, so that $T/_cB = 1$, the only subforest of $T/_cB$ is the empty one.

We map $(A,B)$ to $(B,C)$ with $C \defequiv A \setminus B$.
This is well defined. Each $a \in A \setminus B$ is a vertex of $T/_cB$: if $a < b \leq a'$ for some $b \in B$ and $a' \in A$, then $a' \neq a$ contradicts the fact that $A$ is an antichain, and $a' = a$ gives $a < a$.
It is not a trace leaf of $T/_cB$, since $a \notin B$ and $a$ is not a trace leaf of $T$. A subset of an antichain is an antichain.
(If $A = \{\text{root}\}$ and $B \neq \{\text{root}\}$, then $C$ is the root of $T/_cB$; if $B = \{\text{root}\}$, then $C = \emptyset$.)

Conversely, we map $(B,C)$ to $(A,B)$ with $A \defequiv C \cup \{\, b \in B : b \text{ lies below no } c \in C \,\}$.
This $A$ is an antichain: an element $c \in C$ is a vertex of $T/_cB$, hence does not lie strictly below any $b \in B$, and it is not a trace leaf of $T/_cB$, hence $c \neq b$;
and no retained $b$ lies below a $c$, by definition.
No element of $A$ is a trace leaf of $T$: for $c \in C$ by the identification of vertices above, and for $b \in B$ because $B$ is a subforest.
No element of $A$ is the root of $T$, except in two cases, in both of which $A = \{\text{root}\}$:
if $C$ is the root of $T/_cB$, every $b \in B$ lies below it, so $A = C$; and if $B = \{\text{root}\}$, then $C = \emptyset$ and $A = B$.
Hence $A$ is a subforest of $T$. Every $b \in B$ satisfies $b \leq a$ for some $a \in A$ by construction.

The two maps are inverse to each other. Starting from $(B,C)$, we have $A \setminus B = C$, because the vertices added to $C$ belong to $B$ and $C \cap B = \emptyset$.
Starting from $(A,B)$ and putting $C = A \setminus B$: if $b \in B$ lies below no element of $A \setminus B$, then the $a \in A$ with $b \leq a$ belongs to $A \cap B$, and $b = a$ since $B$ is an antichain;
conversely, an element of $A \cap B$ lies below no other element of $A$. Hence $C \cup \{\, b \in B : b \text{ lies below no } c \in C \,\} = A$.
We thus have a bijection between the terms of the two sides.

\paragraph{Agreement of the factors.}
We compare the factors of corresponding terms.
The first factor is $F_B$ on both sides.
The second factor is $\bigsqcup_{a \in A} T_a/_c(B \cap T_a)$ on the left, where the components with $a \in B$ are $1$, and $\bigsqcup_{c \in C} (T/_cB)_c$ on the right.
For $c \in A \setminus B$ the two are the same tree, and each trace leaf at a vertex $b \in B$ carries the label $\lambda(T_b)$ on both sides. The second factors therefore agree, whatever the rule $\lambda$ is.
If $A = \{\text{root}\}$, the third factor is $1$ on both sides. Otherwise the third factor has, on both sides, the shape of $T/_cA$. At a vertex $a \in A \cap B$, the label is $\lambda(T_a)$ on both sides.
At a vertex $c \in A \setminus B$, the label is $\lambda(T_c)$ on the left and $\lambda(T_c/_c(B \cap T_c))$ on the right, where $B \cap T_c$ consists of accessible vertices of $T_c$.
This is the only place where the two sides can differ.

\paragraph{Proof of Proposition \ref{lem:label-invariance}.}
Suppose that $\lambda(X/_cF) = \lambda(X)$ for every tree $X$ and every subforest $F$ of accessible terms of $X$. Then the labels at the vertices $c \in A \setminus B$ agree as well,
so corresponding terms are equal, and coassociativity follows from the bijection.

Conversely, suppose that $\cop_c$ is coassociative, and let $X$ be a tree and $B_0$ a nonempty antichain of accessible vertices of $X$, with $F \defequiv F_{B_0}$ the corresponding subforest and $X/_cF \defequiv X/_cB_0$
(if the subforest is empty there is nothing to prove). In particular $X$ is not a single leaf.
Pick $\delta \in \SOz$, put $T \defequiv \{\delta, X\}$, and let $w$ be the root of $X$. We compare the coefficients of the basis element
\[
F \ten X/_cF \ten \{\delta, \tr{\lambda(X)}\}
\]
on the two sides.
On either side the third factor has the shape of $T/_cA$. Since $X$ has at least two leaves, $T/_cA$ has exactly two leaves only if $A = \{w\}$ or $A = \{\delta, w\}$, and in the latter case it has no lexical leaf.
Hence only the terms with $A = \{w\}$ contribute.
On the left-hand side these terms are $F_B \ten X/_cB \ten \{\delta,\tr{\lambda(X)}\}$, with $B$ ranging over the subforests of $X$.
Let $N$ be the number of antichains $B$ of accessible vertices of $X$ such that $F_B \cong F$ and $X/_cB \cong X/_cF$; then $N \geq 1$ (as $B_0$ is one of them), and the coefficient on the left-hand side is $N$.
On the right-hand side, the terms with $A = \{w\}$ and a nonempty second factor are $F_B \ten X/_cB \ten \{\delta,\tr{\lambda(X/_cB)}\}$, with $B$ ranging over the same antichains.
For each of the $N$ antichains above, $X/_cB \cong X/_cF$ and hence $\lambda(X/_cB) = \lambda(X/_cF)$. The coefficient on the right-hand side is therefore $N$ if $\lambda(X/_cF) = \lambda(X)$, and $0$ otherwise.
Since $N \geq 1$, coassociativity forces $\lambda(X/_cF) = \lambda(X)$. \qed

\paragraph{Proof of Proposition \ref{prop:resolved}.}
The rule is $\lambda = \tr{\mathrm{res}(\cdot)}$. In $T_c/_c(B \cap T_c)$, the trace leaf at $b \in B$ carries the label $\tr{\mathrm{res}(T_b)}$, and $\mathrm{res}$ restores it to $\mathrm{res}(T_b)$.
On the other hand, $\mathrm{res}(T_c)$ is obtained from $T_c$ by replacing each $T_b$ inside it with $\mathrm{res}(T_b)$ and restoring the remaining trace leaves.
Hence $\mathrm{res}(T_c/_c(B \cap T_c)) = \mathrm{res}(T_c)$, the condition of Proposition~\ref{lem:label-invariance} holds, and $\cop_c$ is coassociative.
The counit axiom also holds: the only term of $\cop_c(T)$ whose left factor is $1$ is $1 \ten T$, and the only term whose right factor is $1$ is $T \ten 1$, since $T/_cA$ retains the root unless $A = \{\text{root}\}$; and $\cou$ is multiplicative by definition. \qed

If trace leaves were counted among the accessible terms, then, for any tree $T$ without trace leaves that is not a single leaf, terms whose second factor has an isolated trace leaf as a component would appear on the right-hand side only,
and coassociativity would fail under every extension.

\subsection{Proposition \ref{prop:grading} (grading and connectedness)}
\label{sec:proofs-grading}

The degree does not depend on the labels of trace leaves, so the argument below is the same for $\cop_c$ with the restoring extension, for the label-forgetting $\copb_c$, and for $\cop_c^{h}$ under the reading of \S\ref{sec:delta-c}. We write it for $\cop_c$ and $\HA$.
A binary tree $T$ with $n$ non-trace leaves and $k$ trace leaves has $n+k$ leaves, $n+k-1$ internal vertices, and $2(n+k)-2$ edges, so
\[
\deg(T) = 2(n+k)-2 + n = 3n + 2k - 2.
\]
For forests, take the sum over the components.

\paragraph{Compatibility of the grading with $\cop_c$.}
Suppose a subforest $F_v = T_{v_1} \sqcup \cdots \sqcup T_{v_j}$ with $j$ components is extracted from $T$.
If $F_v$ has $n_v$ non-trace leaves and $k_v$ trace leaves, then $\deg(F_v) = 3n_v + 2k_v - 2j$.
The contraction quotient $T/_cF_v$ is the binary tree obtained by replacing each $T_{v_i}$ with a trace leaf, so it has $n - n_v$ non-trace leaves and $k - k_v + j$ trace leaves,
and $\deg(T/_cF_v) = 3(n-n_v) + 2(k-k_v+j) - 2$. The sum equals $3n + 2k - 2 = \deg(T)$.
The terms $T \ten 1$ and $1 \ten T$ are obviously compatible as well. For forests the same holds, since the coproduct is extended multiplicatively.
Hence $\cop_c$ preserves the degree, and that $\sqcup$ adds degrees is clear from the definition.

\paragraph{Elements of degree $0$.}
The only trees with $3n + 2k - 2 = 0$ are those with $n = 0, k = 1$, that is, isolated trace leaves $\tr{S}$ with $S$ arbitrary (a lexical item, with $n = 1, k = 0$, has degree $1$).
Hence the degree-$0$ part of $\HA$ is spanned by the forests all of whose components are isolated trace leaves (including the empty forest $1$).

\paragraph{Closure of the subspace $\HA^{\circ}$.}
Let $\HA^{\circ}$ be the subspace spanned by the forests that have no isolated trace leaf as a component.
Closure under $\sqcup$ is obvious.
As for $\cop_c$: in each term $F_v \ten F/_cF_v$ for $F \in \HA^{\circ}$, the components of the left factor $F_v$ are accessible terms, which are not trace leaves since trace leaves are not counted among the accessible terms,
or whole components of $F$, which are not isolated trace leaves by assumption. Each component of the right factor is $T/_cF_v$ (with $F_v \subsetneq T$ a proper subforest), which is either a rooted binary tree with at least two leaves or,
when $F_v = \emptyset$, $T$ itself, or else $1$, when the whole component $T$ is extracted. None of these is an isolated trace leaf.
Hence $\cop_c(\HA^{\circ}) \subset \HA^{\circ} \ten \HA^{\circ}$, and $\HA^{\circ}$ is a sub-bialgebra.

\paragraph{Conclusion.}
Since the degree-$0$ part of $\HA^{\circ}$ is $\Set{Q}\cdot 1$ only, $(\HA^{\circ}, \sqcup, \cop_c, \deg)$ is a commutative graded connected bialgebra (coassociativity is Proposition~\ref{prop:resolved}).
A graded connected bialgebra has an antipode \citep{Takeuchi1971}: in the formula $S = \sum_{k \geq 0} (-1)^k (\idm - u\circ\cou)^{\conv k}$, with $u$ the unit, the terms with $k > n$ vanish on the degree-$n$ part by connectedness,
so the sum is locally finite, and the finite-dimensionality of the homogeneous components is not needed
(they are infinite-dimensional, since the labels $\tr{S}$ of trace leaves do not enter the degree).
Note that, instead of restricting to $\HA^{\circ}$, one may assign degree $1$ to isolated trace leaves without breaking the compatibility
($\cop_c(\tr{S}) = \tr{S} \ten 1 + 1 \ten \tr{S}$, and an isolated trace leaf does not appear as a component of a factor of any other term).
\qed

\subsection{Lemma (multiplicativity of the Bogolyubov recursion)}
\label{sec:proofs-bogolyubov}

\begin{lemma}
\label{lem:multiplicative}
Let $\HA$ be a commutative algebra with a basis of forests (with product $\sqcup$), and $\cop : \HA \to \HA \ten \HA$ an algebra morphism such that,
for every nonempty forest $x$, the left factor $x'$ of each term of the reduced coproduct $\rcop(x) = \cop(x) - x \ten 1 - 1 \ten x = \sum x' \ten x''$ is
a nonempty proper subforest of $x$ ($\cop_c$, $\cop_\rho$, and $\cop_d$ all qualify; neither coassociativity, nor a grading, nor connectedness is assumed).
Let $(\RBalg, \RBop)$ be a commutative Rota--Baxter algebra of weight $-1$, and $\phi : \HA \to \RBalg$ an algebra morphism. Define $\phi_\pm$ by
\[
\phi_-(x) = -\RBop\big(\phi(x) + \textstyle\sum \phi_-(x')\,\phi(x'')\big), \qquad
\phi_+(x) = (1 - \RBop)\big(\phi(x) + \textstyle\sum \phi_-(x')\,\phi(x'')\big)
\]
(with $\phi_\pm(1) = 1$). This recursion is well defined by induction on the number of vertices, and $\phi_\pm$ are algebra morphisms.
\end{lemma}

\begin{proof}
The recursion evaluates $\phi_-$ only on the left factors $x'$, and $x'$ is a proper subforest of $x$, so it has strictly fewer vertices.
Hence the recursion is well defined by induction on the number of vertices.
The proof of multiplicativity follows that of Connes--Kreimer \citep{CK1998} and does not use coassociativity.
Put $\Bog{\phi}(x) = \phi(x) + \sum \phi_-(x')\phi(x'')$ (the Bogolyubov preparation).
For nonempty forests $x, y$, we show $\phi_-(xy) = \phi_-(x)\phi_-(y)$ by induction on the number of vertices of $xy$.
Since $\cop$ is an algebra morphism, writing $\rcop(x) = x' \ten x''$ in Sweedler notation we have
\[
\rcop(xy) = x \ten y + y \ten x + x y' \ten y'' + y' \ten x y'' + x' y \ten x'' + x' \ten x'' y + x' y' \ten x'' y''.
\]
The left factor of each term is a product with fewer vertices than $xy$, so by the induction hypothesis $\phi_-$ is multiplicative on them
(for $\cop_d$ there may be terms whose right factor $x''$ is $1$, but this does not affect the classification).
Substituting this into $\Bog{\phi}(xy) = \phi(xy) + \sum \phi_-((xy)')\,\phi((xy)'')$ and rearranging, we obtain
\[
\Bog{\phi}(xy) = \Bog{\phi}(x)\,\Bog{\phi}(y) + \phi_-(x)\,\Bog{\phi}(y) + \Bog{\phi}(x)\,\phi_-(y)
\]
(the eight terms correspond one-to-one to the expansion of the three products on the right-hand side).
Applying $\phi_-(x) = -\RBop(\Bog{\phi}(x))$ and the Rota--Baxter identity of weight $-1$,
$\RBop(a)\RBop(b) = \RBop(\RBop(a)\,b + a\,\RBop(b) - ab)$, with $a = \Bog{\phi}(x)$ and $b = \Bog{\phi}(y)$, we get
\[
\phi_-(x)\phi_-(y) = \RBop\big(-\phi_-(x)\Bog{\phi}(y) - \Bog{\phi}(x)\phi_-(y) - \Bog{\phi}(x)\Bog{\phi}(y)\big)
 = -\RBop(\Bog{\phi}(xy)) = \phi_-(xy).
\]
The multiplicativity of $\phi_+ = \phi_- \conv \phi$ follows from the fact that, when $\cop$ is an algebra morphism and $\RBalg$ is commutative, the convolution of two algebra morphisms is an algebra morphism
(again without coassociativity).
\end{proof}

What MCB's semantic composition actually uses is the semiring version \mcb{Prop.~3.1.9, pp.~213--214}.
Since there are no additive inverses, the sign disappears from the definition of $\phi_-$, and the factorization takes the form $\phi_+ = \phi_- \conv \phi$.
Since the proof of Lemma~\ref{lem:multiplicative} uses no subtraction, it carries over to semirings as it stands.

\begin{lemma}
\label{lem:multiplicative-semiring}
Let $(\RBalg, \boxplus, \odot, 0, 1)$ be a commutative semiring, and $\RBop : \RBalg \to \RBalg$ an additive map
($\RBop(0) = 0$, $\RBop(a \boxplus b) = \RBop(a) \boxplus \RBop(b)$) satisfying the Rota--Baxter identity of weight $+1$,
\[
\RBop(a) \odot \RBop(b) = \RBop\big(\RBop(a) \odot b \;\boxplus\; a \odot \RBop(b) \;\boxplus\; a \odot b\big).
\]
Let $\HA$ be as in Lemma~\ref{lem:multiplicative}, and assume moreover that the structure constants of $\cop$ are nonnegative integers
(with respect to the basis of forests, $\cop$ is a sum of basis elements with nonnegative integer coefficients); this holds by definition for $\cop_c$ and $\cop_d$.
Let $\Hsemi$ be the $\Set{N}$-semimodule spanned by the basis of forests, and $\phi : \Hsemi \to \RBalg$ a semiring character
(additive, with $\phi(xy) = \phi(x) \odot \phi(y)$ and $\phi(1) = 1$).
For a nonempty forest $x$, define $\phi_\pm$ by
\[
\Bog{\phi}(x) = \phi(x) \boxplus \textstyle\bigboxplus \phi_-(x') \odot \phi(x''), \qquad
\phi_-(x) = \RBop(\Bog{\phi}(x)), \qquad
\phi_+(x) = (\phi_- \conv \phi)(x) = \Bog{\phi}(x) \boxplus \phi_-(x)
\]
(with $\phi_\pm(1) = 1$). Then $\phi_\pm$ are semiring characters.
Idempotency of addition is not needed.
\end{lemma}

\begin{proof}
The expansion in the proof of Lemma~\ref{lem:multiplicative},
$\Bog{\phi}(xy) = \Bog{\phi}(x) \odot \Bog{\phi}(y) \boxplus \phi_-(x) \odot \Bog{\phi}(y) \boxplus \Bog{\phi}(x) \odot \phi_-(y)$,
is an identity in which the eight terms obtained by rewriting the seven kinds of terms of $\rcop(xy)$ by the induction hypothesis correspond one-to-one
(under the commutativity of $\RBalg$) to the eight terms of the expansion of the three products on the right-hand side; it uses neither subtraction nor the collapsing of repeated terms.
Since the structure constants of $\cop$ are nonnegative integers, $\cop(xy) = \cop(x)\cop(y)$ holds coefficientwise as an identity in the free $\Set{N}$-semimodule,
and no subtraction is needed to define $\rcop(xy)$ as ``the terms other than $xy \ten 1$ and $1 \ten xy$''
(among the nine products in $\cop(x)\cop(y)$, only $(x \ten 1)(y \ten 1)$ and $(1 \ten x)(1 \ten y)$ yield $xy \ten 1$ and $1 \ten xy$).
We extend $\Bog{\phi}$ and $\phi_-$ additively to sums with $\Set{N}$-coefficients.
Applying the identity of weight $+1$ with $a = \Bog{\phi}(x)$ and $b = \Bog{\phi}(y)$ then gives
\[
\phi_-(x) \odot \phi_-(y)
 = \RBop\big(\phi_-(x) \odot \Bog{\phi}(y) \boxplus \Bog{\phi}(x) \odot \phi_-(y) \boxplus \Bog{\phi}(x) \odot \Bog{\phi}(y)\big)
 = \RBop(\Bog{\phi}(xy)) = \phi_-(xy).
\]
The multiplicativity of $\phi_+ = \phi_- \conv \phi$ follows as in Lemma~\ref{lem:multiplicative}, from the fact that $\cop(xy) = \cop(x)\cop(y)$ holds as an identity with $\Set{N}$-coefficients
and from the commutativity of $\RBalg$.
\end{proof}

\begin{remark}
\label{rem:semiring-examples}
MCB states, as a separate case, that $\phi_\pm$ are multiplicative also when $\RBop$ is a Rota--Baxter operator of weight $-1$ on a semiring, under the additional condition
\begin{equation}
\label{eq:mcb318}
\RBop(a \odot b) \boxplus \RBop(a) \odot \RBop(b) = \RBop(a) \odot \RBop(b)
\end{equation}
\mcb{Remark 3.1.10, (3.1.8), p.~214} (the source is \citealp{MT2015}).
This is not a separate case, however, but a corollary of Lemma~\ref{lem:multiplicative-semiring}.
In a semiring, the identity of weight $-1$ takes the form (there being no subtraction) $\RBop(a)\odot\RBop(b) \boxplus \RBop(a \odot b) = \RBop(a \odot \RBop(b)) \boxplus \RBop(\RBop(a) \odot b)$,
and by \eqref{eq:mcb318} its left-hand side equals $\RBop(a)\odot\RBop(b)$.
Then, by the additivity of $\RBop$, the right-hand side of the identity of weight $+1$ becomes
$\RBop(a \odot \RBop(b)) \boxplus \RBop(\RBop(a) \odot b) \boxplus \RBop(a \odot b) = \RBop(a)\odot\RBop(b) \boxplus \RBop(a \odot b) = \RBop(a)\odot\RBop(b)$
(using \eqref{eq:mcb318} once more at the end), so $\RBop$ satisfies the identity of weight $+1$.
That is, weight $-1$ together with \eqref{eq:mcb318} implies weight $+1$, and in neither case is idempotency of addition needed.
One can also check directly that the threshold maps $c_\lambda$ on the Viterbi semiring $([0,1], \max, \cdot)$ used in MCB's second toy model \mcb{Lemma 3.2.11, p.~228} satisfy the identity of weight $+1$
for $0 < \lambda \le 1$ (MCB states the lemma for $\lambda \in [0,1]$, but $c_0$ sends $0$ to $1$ and is therefore not additive).
\end{remark}

\subsection{Proposition \ref{prop:quasi-hopf} (quasi-Hopf structures over a commutative algebra)}
\label{sec:proofs-quasi-hopf}

In a Drinfeld quasi-bialgebra \citep{Drinfeld1990}, coassociativity is replaced by the requirement that there exist an invertible element $\Phi \in \HA \ten \HA \ten \HA$ such that
\[
(\idm \ten \cop)\circ\cop(h) = \Phi \cdot \big((\cop \ten \idm)\circ\cop(h)\big) \cdot \Phi^{-1}
\]
holds for all $h$ (together with the pentagon condition on $\Phi$).
If $\HA$ is commutative, then so is $\HA \ten \HA \ten \HA$, and the right-hand side equals $(\cop \ten \idm)\circ\cop(h)$.
That is, coassociativity holds strictly. \qed

\nocite{MCB2025}
\bibliographystyle{plainnat}
\bibliography{refs}

\end{document}